\documentclass[10pt,]{nvidiatechreport}

\usepackage{mdframed}
\usepackage[utf8]{inputenc} % allow utf-8 input
\usepackage[T1]{fontenc}    % use 8-bit T1 fonts

\usepackage{amsfonts}       % blackboard math symbols
\usepackage{nicefrac}       % compact symbols for 1/2, etc.
\usepackage{microtype}      % microtypography
\usepackage[dvipsnames]{xcolor}         % colors
\usepackage{multirow}
\usepackage{multicol}
\usepackage{graphicx}
\usepackage[numbers]{natbib}
\usepackage{tabto}
\usepackage{xspace}
\usepackage{amsmath}
\usepackage{adjustbox}
\usepackage{enumitem}
\usepackage{wrapfig}
\usepackage{dblfloatfix}

\usepackage{footmisc}
\usepackage{listings}

\usepackage{float}

\usepackage{hyperref}
\usepackage{array}
\usepackage{ragged2e}
\usepackage{subcaption}
\usepackage{caption}

\usepackage{natbib}
\usepackage{amsmath}
\usepackage{amssymb}
\usepackage{mathtools}
\usepackage{amsthm}
\usepackage{enumitem}
\setlist[itemize]{itemsep=0.5em}
\usepackage{multirow}
\usepackage{booktabs}

\usepackage[capitalize,noabbrev]{cleveref}

\usepackage{algorithm}
\usepackage{algorithmicx}
\usepackage{algpseudocode}
\usepackage{xspace}
\usepackage{csquotes}
\usepackage{listings}
\usepackage{xcolor}
\definecolor{NvidiaGreen}{rgb}{0,0.392,0}
\newtheorem{definition}{Definition}[section]

\newcommand{\tabsmall}{\scriptsize}

\usepackage{multirow}
\usepackage{multicol}
\usepackage{enumitem}
\usepackage[normalem]{ulem}
\usepackage{pifont}
\usepackage{tikz}
\usetikzlibrary{calc}

\usepackage[most]{tcolorbox}

\definecolor{lightgray}{gray}{0.95} 
\definecolor{darkblue}{rgb}{0,0,0.6} 
\definecolor{nvgreen}{cmyk}{50, 0, 100, 0}

\newcommand{\name}{PRESTO }

\renewcommand{\footrulewidth}{\iffootnote 0.6pt\else 0pt\fi}

\title{PRESTO: Prefix-Aligned Tree Drafting for Diffusion Speculative Decoding}

\author{
\textbf{Zheng Wang}$^{1,*}$ \quad
\textbf{Zhifan Ye}$^{2,3,*}$ \quad
\textbf{Qi Cheng}$^{3}$ \quad
\textbf{Yonggan Fu}$^{3}$ \quad
\textbf{Ziyan Wang}$^{2}$ \quad
\textbf{Feng Zhu}$^{2}$ \\
\textbf{Haozhe Zhao}$^{1}$ \quad
\textbf{Jan Kautz}$^{3}$ \quad
\textbf{Pavlo Molchanov}$^{3}$ \quad
\textbf{Humphrey Shi}$^{2,3}$ \quad
\textbf{Minjia Zhang}$^{1}$ \\
\mbox{}\\
$^{1}$University of Illinois Urbana-Champaign \quad
$^{2}$Georgia Institute of Technology \quad
$^{3}$NVIDIA \\
$^{*}$Equal contribution
}

\begin{abstract}
\textbf{Abstract:} Diffusion Large Language Models (dLLMs) have recently emerged as a 
promising alternative to autoregressive (AR) LLMs, offering parallel token 
generation. Recent works have shown that dLLMs are particularly well-suited as draft models for speculative decoding (SD), as they can efficiently generate an entire block of draft tokens in parallel within a single forward pass. However, existing diffusion-based drafting methods primarily rely on linear drafting (single-path), despite diffusion models simultaneously 
producing multiple candidate tokens at multiple positions that naturally 
induce a large combinatorial space of possible decoding paths. As a result, these methods can only explore a tiny fraction of the available candidate space, inherently limiting the achievable acceptance length and decoding efficiency. To fully utilize this rich multi-candidate structure, we apply tree-based drafting to diffusion drafters, enabling the exploration of diverse candidate paths. Nevertheless, we find that applying naive tree-based drafting is suboptimal due to a fundamental mismatch between diffusion draft confidence and prefix-based AR verification: diffusion marginals are inherently prefix-blind, which can lead to unreliable path ranking. We propose PRESTO, a principled framework that extends tree-based drafting to diffusion drafters while resolving the fundamental mismatch between diffusion draft confidence and prefix-based AR verification through \emph{\textbf{PRE}fix-aligned \textbf{S}coring} and \emph{priority-based \textbf{T}ree search} for diffusion speculative dec\textbf{O}ding. The key principles behind PRESTO are that (1) candidate ranking should align with the prefix-based nature of AR verification, and (2) tree construction should prioritize candidate paths with high verification potential to maximize acceptance length. We also demonstrate that PRESTO is designed to be a general tree drafting framework applicable to both dedicated diffusion drafter SD and self-speculative dLLMs. Extensive experiments show that PRESTO achieves an average of \textbf{1.5$\times$} end-to-end throughput speedup on the state-of-the-art dedicated diffusion drafter SD and an average of \textbf{1.12$\times$} on self-speculative diffusion LLMs across diverse benchmarks.
\end{abstract}

\date{}
\begin{document}
\maketitle

\begin{figure}[h]
    % \vspace{-0.5em}
    \centering
    \includegraphics[width=\linewidth]{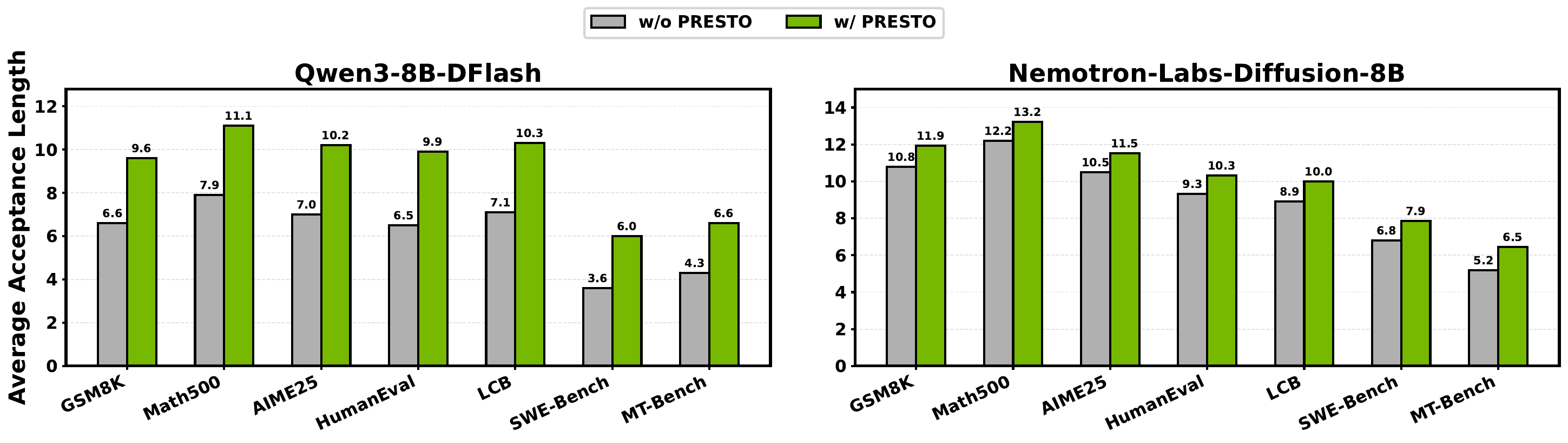}
    \vspace{-1.5em}
    \caption{Acceptance length comparison of PRESTO on diffusion-based SD, including dedicated diffusion drafter SD (Qwen3-8B-DFlash) and self-speculative dLLM (Nemotron-Labs-Diffusion-8B; marginal-only scoring with no prefix-signal) across diverse benchmarks. PRESTO consistently improves acceptance length over the baseline across all tasks, enabling more tokens to be accepted in each verification round.}
    \label{fig:observations}
    \vspace{-0.5em}
\end{figure}

\section{Introduction}

Diffusion Large Language Models (dLLMs) have recently emerged as a 
promising alternative to autoregressive (AR) LLMs, introducing a new paradigm for multi-token generation~\cite{nie2025largelanguagediffusionmodels,ye2025dream7bdiffusionlarge,wu2025fastdllmv2efficientblockdiffusion,bie2026llada21speedingtextdiffusion,cheng2025sdarsynergisticdiffusionautoregressionparadigm,fu2026efficientdlmautoregressivediffusionlanguage}. However, achieving high-quality 
generation in practice typically requires 
iterative denoising procedures and carefully designed token unmasking 
strategies, which introduce additional computation and partially offset 
the efficiency gains~\cite{wu2025fastdllmv2efficientblockdiffusion,wu2025fastdllmtrainingfreeaccelerationdiffusion}. Interestingly, recent works have shown that diffusion models are particularly well-suited as draft models in speculative decoding~\cite{liu2025tidarthinkdiffusiontalk,li2025diffuspecunlockingdiffusionlanguage,chen2026dflashblockdiffusionflash,yu2026introspectivediffusionlanguagemodels}. Their ability to generate multiple candidate tokens in parallel naturally aligns with the proposal stage of speculative decoding (SD), enabling highly efficient draft token generation. This insight has led to a growing line of work that leverages diffusion models as effective drafters, including dedicated diffusion-based drafter for AR LLMs (e.g., dFlash~\cite{chen2026dflashblockdiffusionflash}) and self-speculative dLLMs (e.g., TiDAR~\cite{liu2025tidarthinkdiffusiontalk}, I-DLM~\cite{yu2026introspectivediffusionlanguagemodels}, Nemotron-Labs-Diffusion~\cite{fu2026nemotron}). For example, dFlash demonstrates that well trained lightweight diffusion drafters can substantially accelerate speculative decoding through efficient parallel draft token generation while maintaining high draft quality, highlighting the practical potential of diffusion-based drafting.

\noindent However, existing diffusion-based drafting methods primarily rely on 
\emph{linear (single-path) drafting}, where tokens are generated along 
a single trajectory and verified sequentially. While this design is 
effective in AR settings, it is inherently suboptimal for 
diffusion models. Due to the non-autoregressive nature of diffusion generation, multiple plausible candidates can exist simultaneously, while the decoding path with the highest acceptance length may not correspond to the top-1 marginal trajectory. Empirically, we observe that even strong diffusion drafters in dFlash yield only modest accepted lengths under single-path drafting (See Figure~\ref{fig:observations}(a)), indicating that a significant portion of the proposal space remains unexplored.

\noindent A natural direction to address this limitation is to move beyond linear drafting and explore multiple candidate paths jointly. In AR settings, prior work has shown that introducing tree-based drafting can substantially increase acceptance length by expanding multiple candidate paths in parallel, thereby improving the utilization of each verification step~\cite{Miao_2024,cai2024medusasimplellminference,li2024eagle2fasterinferencelanguage,li2025eagle3scalinginferenceacceleration,svirschevski2024specexecmassivelyparallelspeculative,chen2025sequoiascalablerobusthardwareaware}. In particular, since the target model is often not fully compute-bound during verification~\cite{liu2025tidarthinkdiffusiontalk} for small batch size cases, allocating additional drafting compute to explore a richer candidate set can lead to more tokens being accepted per step. The success of tree-based drafting in AR settings raises a key research question:

\emph{Can we leverage tree-based drafting in diffusion drafters to further improve the end-to-end throughput of diffusion-based speculative decoding by increasing acceptance length?}

\noindent While this direction is promising, we find that naively applying tree-based drafting to diffusion models is often suboptimal. 
Diffusion drafters generate position-wise token distributions in parallel, without strict prefix dependencies, which enables efficient and flexible construction of tree-structured candidates. 
However, this same property also implies that the resulting scores are marginal token probabilities, rather than prefix-conditioned likelihoods. 
Consequently, these scores cannot accurately reflect how earlier token choices influence downstream acceptance (e.g., a token that is likely in isolation may become unlikely once conditioned on the chosen prefix), creating a fundamental mismatch between draft scoring and prefix-based AR verification.

\noindent In this work, we propose PRESTO, a principled tree-based diffusion drafting framework via \emph{prefix-aligned scoring} and \emph{priority-based 
tree search}. The key principles behind PRESTO are that (1) candidate ranking should align with the prefix-based nature of AR verification, and (2) tree construction should prioritize candidate paths with high verification potential to maximize acceptance length. Specifically, we minimally adjust the 
diffusion marginal by incorporating a prefix-dependent correction, 
so that the resulting score better reflects the acceptance behavior 
under sequential verification. This score is then used to guide a 
priority-based expansion strategy, which allocates computation to 
high-quality candidate paths during tree construction to maximize acceptance length. 
As a result, PRESTO enables effective tree-based diffusion drafting and 
significantly improves acceptance length and throughput on both dedicated diffusion-based drafters for AR LLMs and self-speculative dLLMs. Specifically, we summarize our contributions as follows:
\begin{itemize}

\item \textbf{We identify a fundamental mismatch between diffusion-based 
draft scoring and prefix-based AR verification.} 
We show that while diffusion probabilities provide well-calibrated marginal acceptance signals, they do not fully capture the prefix-conditioned nature of AR verification, which can lead to suboptimal path ranking during tree construction and ultimately limit acceptance length.

\item \textbf{We design prefix-aligned scoring to address the above mismatch.} 
We formulate a principled, minimal correction to diffusion marginals by incorporating prefix-dependent signals, yielding a scoring function that is both faithful to the diffusion drafters and compatible with prefix-based verification.

\item \textbf{We formalize and enable tree-based speculative drafting for diffusion drafters.}
Building upon the proposed prefix-aligned scoring, we introduce PRESTO, a principled tree-based diffusion drafting framework that efficiently explores high-quality draft paths through priority-based tree expansion, substantially improving acceptance length and decoding throughput.

\item \textbf{PRESTO generalizes across both dedicated and self-speculative diffusion SD frameworks.} 
We further extend PRESTO to self-speculative dLLMs as a drop-in replacement for linear drafting in the parallelized verify-and-draft pipeline, making PRESTO the first tree-based drafting framework applicable to both dedicated diffusion drafters and self-speculative diffusion LLMs. Across diverse benchmarks and model sizes, PRESTO achieves an average of \textbf{1.5$\times$} throughput speedup on dFlash and an average of \textbf{1.12$\times$} throughput speedup on Nemotron-Labs-Diffusion over linear drafting baselines.

\end{itemize}

\section{Preliminaries}
\label{sec:preliminaries}
\subsection{Tree-based Speculative Decoding}
\label{sec:prelim_tree}

Speculative decoding accelerates autoregressive generation of a target 
LLM $p_T$ by leveraging a lightweight draft model. To improve 
acceptance, the draft model can construct a token tree encoding multiple candidate sequences~\citep{Miao_2024,sun2024spectrfastspeculativedecoding,chen2025sequoiascalablerobusthardwareaware}. 
% \minjia{Probably need a reference.} \zheng{I added works that exploring token tree} 
The target LLM verifies 
candidates in a prefix-based manner: starting from the root, 
it accepts tokens sequentially until a mismatch occurs.

\noindent\textbf{Acceptance factorization.}
For a candidate path $P = (x_1, \ldots, x_k)$, let $a_i \in \{0, 1\}$ 
denote the indicator that the $i$-th token is accepted. Acceptance 
proceeds left-to-right, so by the chain rule
\begin{equation}
\Pr(P \text{ accepted}) 
= \prod_{i=1}^{k} \Pr\!\bigl(a_i = 1 \,\bigm|\, a_{<i} = 1,\, x_{\leq i}\bigr),
\label{eq:acceptance}
\end{equation}
where $a_{<i} = 1$ denotes the event that all earlier tokens have 
been accepted. Each factor depends on the realized prefix $x_{<i}$, 
making verification intrinsically prefix-conditional.

\noindent\textbf{Tree construction objective.}
Let $\mathcal{B}_B = \{\mathcal{T} : |\mathcal{T}| \le B\}$ denote token trees with at most $B$ nodes where $\mathcal{T}$ is the tree budget, and let $\alpha_{\mathcal{T}}(P)$ denote the longest prefix of $P$ contained in $\mathcal{T}$. The ideal objective is
\begin{equation}
\mathcal{T}^* \in \arg\max_{\mathcal{T}\in \mathcal{B}_B} \mathbb{E}_{P\sim p_T}[\alpha_{\mathcal{T}}(P)].
\end{equation}
Direct optimization is infeasible, as evaluating $p_T$ requires target-model forward passes. We use a surrogate distribution $\tilde{p}$ over candidate paths. A standard decomposition (Appendix~\ref{appendix:decomposition}) yields
\begin{equation}
\mathbb{E}_{P\sim \tilde{p}}[\alpha_{\mathcal{T}}(P)]
= \sum_{u\in \mathcal{T}} \tilde{p}(u),
\label{eq:surrogate-objective}
\end{equation}
where $\tilde{p}(u)$ denotes the probability that prefix $u$ matches the sampled path. This reduces tree construction to selecting prefixes with large probability mass. Since $\tilde{p}(u)$ factorizes along the prefix,
\begin{equation}
\log \tilde{p}(u) = \sum_{i=1}^{|u|} \log \tilde{p}(u_i \mid u_{<i}),
\label{eq:log-prefix-prob}
\end{equation}
providing an additive scoring rule that enables priority-based expansion.

\subsection{Prefix-Aligned Surrogate Distributions}
\label{sec:prelim_faithful}

\begin{definition}[Prefix-Aligned Surrogate]
\label{def:prefix-faithful}
A surrogate distribution $\tilde{p}$ over $\mathcal{V}^k$ is \emph{prefix-aligned} if there exists some position $i \ge 2$ such that the conditional distribution $\tilde{p}(u_i \mid u_{<i})$ is not invariant to the prefix, i.e., there exist $u_{<i} \neq u'_{<i}$ with
\begin{equation}
\tilde{p}(u_i \mid u_{<i}) \neq \tilde{p}(u_i \mid u'_{<i}).
\label{eq:nontriviality}
\end{equation}

\end{definition}

\noindent The nontriviality clause is essential. Under an independence assumption, any joint distribution admits a trivial prefix-blind factorization of the form $\tilde{p}(u) = \prod_i \tilde{p}_i(u_i)$, where each token is scored independently of the prefix. Such decompositions assign identical per-position contributions regardless of the realized prefix and therefore fail to capture the prefix-conditioned structure in Eq.~\eqref{eq:acceptance}.

\subsection{Autoregressive vs.\ Diffusion Surrogate Distributions}
\label{sec:prelim_contrast}
Existing tree drafting methods for AR drafters implicitly assume autoregressive prefix-conditioned scores~\citep{Miao_2024,li2024eagle2fasterinferencelanguage,sun2024spectrfastspeculativedecoding,chen2025sequoiascalablerobusthardwareaware}. 
% \minjia{Need citations, otherwise it is unclera what "existing tree drafting methods" refer to after a year this paper is published.} \zheng{I added works that use AR conditional score to construct tree} 
This assumption fundamentally breaks under diffusion drafting. 

\noindent\textbf{Autoregressive drafting yields a prefix-aligned surrogate.}
An AR draft model $p_D$ defines a prefix-conditioned 
distribution over sequences via the chain rule:
\begin{equation}
\tilde{p}_{\text{AR}}(P) := p_D(P) 
= \prod_{i=1}^{k} p_D(x_i \mid x_{<i}).
\label{eq:ar-surrogate}
\end{equation}
Each conditional $p_D(x_i \mid x_{<i})$ is a genuine function of the 
prefix, a well-trained AR drafter approximates the target's 
prefix conditionals, so $\tilde{p}_{\text{AR}}$ is prefix-aligned 
(Definition~\ref{def:prefix-faithful}) and its log-probability 
factorization in Eq.~\eqref{eq:ar-surrogate} provides an effective 
surrogate score for tree construction.

\noindent\textbf{Diffusion drafting yields a prefix-blind marginal surrogate.}
A diffusion drafter produces a collection 
of \emph{position-wise marginals} $q_1, \ldots, q_k$ over $\mathcal{V}$. 
Each $q_i(x_i)$ is the marginal distribution of position $i$; it does not condition on the prefix $x_{<i} = (x_1, \ldots, x_{i-1})$.\footnote{The 
dependence on the prompt and previously generated context, encoded 
in a shared latent state, is suppressed in this section for notational 
clarity.} The induced 
factorized distribution over candidate paths is
\begin{equation}
\tilde{p}_{\text{diff}}(P) := \prod_{i=1}^{k} q_i(x_i).
\label{eq:marginal-surrogate}
\end{equation}

\noindent The marginal surrogate $\tilde{p}_{\text{diff}}$ is not 
prefix-aligned: each conditional 
$\tilde{p}_{\text{diff}}(x_i \mid x_{<i}) = q_i(x_i)$ takes the same 
value regardless of $x_{<i}$, violating the nontriviality clause of 
Definition~\ref{def:prefix-faithful}. The structural mismatch with 
prefix-based verification is \emph{intrinsic} to the marginal factorization, 
not an artifact of any particular drafter or training procedure. In Section~\ref{sec:Observations}, we will show that while this mismatch does not preclude diffusion-based drafting from being effective, it can affect the quality of path ranking.

\section{Observations}
\label{sec:Observations}
\begin{figure}[h]
    \vspace{-0.5em}
    \centering
    \includegraphics[width=\linewidth]{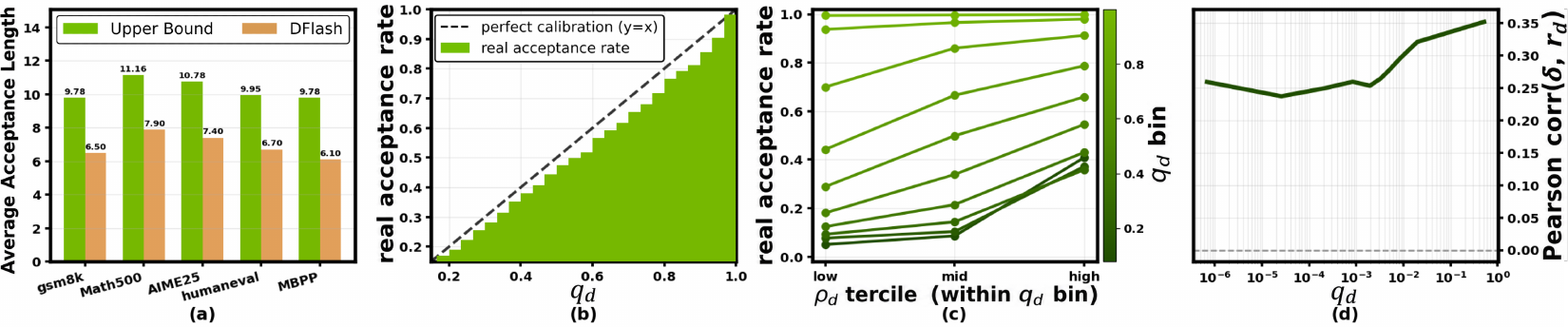}
    \vspace{-1.5em}
    \caption{(a) Linear drafting leaves room for improvement via multi-path exploration.
(b) The predicted confidence from $q_d$ closely matches the empirical acceptance rate. (c) Within fixed $q_d$ bins, empirical acceptance rates increase from low to high $\rho_d$ terciles, indicating that prefix-conditioned compatibility captures predictive signal beyond the marginal confidence $q_d$. (d) $\rho_d$ provides a correction direction aligned with the error of $q_d$, although it remains noisy (correlation $< 1$). 
}
    \label{fig:observations}
    \vspace{-0.5em}
\end{figure}

\noindent In this section, we use Qwen3-4B-DFlash with mask block size being 16 as an example to analyze the characteristics of diffusion-based draft scoring. To study the impact of tree construction, we adopt the tree expansion strategy from EAGLE~\cite{li2024eagle2fasterinferencelanguage,li2025eagle3scalinginferenceacceleration} as the default tree construction algorithm. For each candidate token, we record its diffusion draft probability $q_d$, prefix-conditioned score $\rho_d$, and final acceptance outcome. The prefix-conditioned score $\rho_d$ is instantiated using an $n$-gram model to capture compatibility with prefix-based AR verification. 
Unless otherwise specified, results are aggregated across datasets.

\noindent\textbf{Observation 1: Linear drafting leaves substantial room for improvement through multi-path exploration.} We first compare the acceptance length achieved by linear drafting with an oracle upper bound equivalent to exhaustive multi-path exploration. In our setting, the diffusion drafter generates a block of $16$ draft positions simultaneously, naturally exposing a rich multi-candidate draft space within a single forward pass. To estimate the achievable upper bound, at each position, we retain the top-10 candidates from the diffusion marginal distribution and perform a position-wise oracle matching analysis. Specifically, we check whether the ground-truth token at each position appears in the corresponding top-10 candidate set, and then compute the longest contiguous prefix length covered by these matches.\footnote{The reported upper bound is equivalent to the result that would be obtained by exhaustive enumeration. However, we do not explicitly enumerate all $10^{16}$ paths. Instead, we exploit the fact that acceptance is determined by whether the ground-truth token is covered by the top-k candidate set at each position. This allows us to compute the same oracle upper bound efficiently through position-wise matching.}
As shown in Figure~\ref{fig:observations}(a), a substantial gap consistently exists between linear drafting and this upper bound across all tasks. For example, on GSM8K the acceptance length improves from $6.5$ to $9.78$, while on Math500 it increases from $7.9$ to $11.16$. Similar trends are observed on AIME25, HumanEval, and MBPP. These results suggest that existing linear drafting strategies fail to fully exploit the rich combinatorial candidate space already produced by diffusion drafters, leaving significant room for improvement via multi-path exploration.

\noindent\textbf{Observation 2: Marginal confidence correlates with overall acceptance.} We further examine whether the diffusion drafter provides informative estimates of downstream verification success. Specifically, conditioned on the preceding prefix being accepted by the verifier, we compute the diffusion marginal confidence $q_d$ for the next draft token using the exponentiated marginal probability under the diffusion distribution, and compare it against the empirical acceptance rate observed during verification. As shown in Figure~\ref{fig:observations}(b), the acceptance rate increases monotonically with $q_d$, demonstrating a strong positive correlation between diffusion confidence and actual verification success. This suggests that the diffusion marginal distribution already captures meaningful global acceptance trends, and that higher-confidence draft tokens are substantially more likely to be accepted. At the same time, the calibration curve consistently remains below the ideal $y=x$ line, indicating that marginal confidence alone is still imperfect and does not fully characterize verifier acceptance. Nevertheless, $q_d$ provides a strong global signal for estimating acceptance likelihood and serves as a useful foundation for candidate ranking.

\noindent\textbf{Observation 3: Marginal signals are insufficient but can be improved by prefix conditioning.}
Despite its strong global correlation with acceptance, $q_d$ alone is insufficient for reliable token-level ranking: as shown in Figure~\ref{fig:observations}(c), tokens with similar $q_d$ can yield substantially different acceptance outcomes when grouped by $\rho_d$. This implies that $q_d$ is not a sufficient statistic for predicting acceptance, and that $\rho_d$ provides complementary, prefix-conditioned information. This additional prefix-conditioned signal also remains directionally informative: Figure~\ref{fig:observations}(d) shows that the correction term $\delta = \log \rho_d - \log q_d$ positively correlates with the true residual $r_q = \log p_T - \log q_d$. This indicates that $\rho_d$ provides a correction direction that aligns with the true error of the draft model. Importantly, the correlation remains significantly below 1, indicating that $\rho_d$ is not a direct proxy for the target distribution. Instead, it acts as a noisy but directionally aligned correction signal.

\noindent\textbf{Insights.}
The draft probability $q_d$ provides informative estimates of global acceptance, but is still insufficient for accurate path ranking. The prefix-conditioned signal $\rho_d$ offers complementary information that aligns with the error of $q_d$, but is itself noisy. 
Together, these observations suggest that \emph{an effective scoring should preserve $q_d$ while incorporating a controlled correction based on $\rho_d$}.

\section{PRESTO}
We now introduce PRESTO, a tree-based diffusion drafting framework that augments diffusion drafter with prefix-aligned information, enabling more effective candidate exploration under prefix-conditioned AR verification. In Section \ref{sec:method_scoring}, we introduce prefix-aligned scoring that addresses the aforementioned structural mismatch. In Section \ref{sec:method_search}, we describe tree drafting via priority-based tree search. In Section \ref{sec.self_speculative}, we elaborate how to extend tree drafting to self-speculative dLLMs.

\subsection{Prefix-Aligned Scoring via Minimal Correction}
\label{sec:method_scoring}

\textbf{From surrogate distribution to scoring function.}
Section~\ref{sec:preliminaries} reduces tree construction to ranking prefixes by $\log \tilde{p}(u)$ (Eq.~\eqref{eq:log-prefix-prob}). For diffusion drafting, the natural marginal surrogate $\tilde{p}_{\text{diff}}(P) = \prod_i q_i(x_i)$ is prefix-blind (Eq.~\eqref{eq:marginal-surrogate}), creating a mismatch with prefix-based verification. We therefore seek a prefix-aligned surrogate $\tilde{p}^\star$ that preserves the calibrated drafter signal while incorporating a prefix-conditioned correction in terms of the observations and insights from Section~\ref{sec:Observations}.

\noindent\textbf{Prefix-aligned surrogate.}
Let $q_d(t)$ denote the diffusion drafter's marginal at 
position $d$, and let $\rho_d(t \mid c_d)$ 
denote a tractable prefix-conditioned signal, where 
$c_d = (t_1, \ldots, t_{d-1})$ is the within-block prefix. We define the prefix-aligned conditional as a 
multiplicative combination:
\begin{equation}
p^\star_d(t \mid c_d) 
\propto q_d(t) \, \rho_d(t \mid c_{d})^{\lambda_d},
\label{eq:prefix-aligned-conditional}
\end{equation}
where $\lambda_d \geq 0$ controls the strength of the prefix 
correction. The induced joint surrogate distribution is
\begin{equation}
\tilde{p}^\star(P) := \prod_{d=1}^{k} p^\star_d(t_d \mid c_d),
\label{eq:joint-surrogate}
\end{equation}
which is prefix-aligned (Definition~\ref{def:prefix-faithful}) 
whenever $\lambda_d > 0$ and $\rho_d$ depends nontrivially on $c_d$. 
Eq.~\eqref{eq:joint-surrogate} thereby provides a candidate 
surrogate that addresses the structural mismatch of 
$\tilde{p}_{\text{diff}}$. This form can also be interpreted as the solution to a KL-regularized objective that balances closeness to $q_d$ with alignment to $\rho_d$ (see Appendix~\ref{app:kl-derivation}).

{For simplicity, we use the unnormalized log-score induced by Eq.~\eqref{eq:prefix-aligned-conditional}. Specifically, we define the token-level score as}
$s_{d,t}(c_d) = \log q_d(t) + \lambda_d \log \rho_d(t \mid c_d)$,
and the path score
\begin{equation}
S(P) = \sum_{d=1}^{k} s_{d,t_d}(c_d), 
\qquad c_d = (t_1, \ldots, t_{d-1}).
\label{eq:path-score}
\end{equation}
Eq.~\eqref{eq:path-score} decomposes additively over depth, 
supporting incremental priority-based tree expansion.

\subsection{Priority-Based Tree Construction}
\label{sec:method_search}

\begin{minipage}[t]{0.493\textwidth}
\vspace{-1.5em}
\begin{algorithm}[H]
\scriptsize
\caption{Beam Search with Global Retention}
\label{alg:beam}
\begin{algorithmic}[1]
\Procedure{BeamSearch}{$s, B, b, W, D$}
    \State $\mathcal{P}\gets\{\text{root}\},\ \mathcal{T}\gets\{\text{root}\}$
    \For{$d=1$ to $D$ \textbf{while} $|\mathcal{T}|<B$}
        \State $\mathcal{P}_{\text{new}}\gets\emptyset$
        \For{each $v\in\mathcal{P}$}
            \State $\mathcal{C}\gets\textsc{Expand}(v,b,s)$
            \State $\mathcal{P}_{\text{new}}\gets\mathcal{P}_{\text{new}}\cup\mathcal{C}$
            \State $\mathcal{T}\gets\mathcal{T}\cup\mathcal{C}$
        \EndFor
        \State $\mathcal{P}\gets\textsc{TopW}(\mathcal{P}_{\text{new}},W)$
    \EndFor
    \State \Return $\textsc{TopB}(\mathcal{T},B)$
\EndProcedure
\end{algorithmic}
\end{algorithm}
\end{minipage}
\hfill
\begin{minipage}[t]{0.493\textwidth}
\vspace{-1.5em}
\begin{algorithm}[H]
\scriptsize
\caption{Best-First Search}
\label{alg:bfs}
\begin{algorithmic}[1]
\Procedure{BestFirst}{$s, B, b$}
    \State $\mathcal{P}\gets\{\text{root}\},\ \mathcal{T}\gets\{\text{root}\}$
    \State $S(\text{root})\gets 0$
    \While{$|\mathcal{T}|<B$}
        \State $v^\star\gets\arg\max_{u\in\mathcal{P}}S(u)$
        \State $\mathcal{P}\gets\mathcal{P}\setminus\{v^\star\}$
        \State $\mathcal{C}\gets\textsc{Expand}(v^\star,b,s)$
        \State Insert $\mathcal{C}$ into priority queue $\mathcal{P}$
        \State $\mathcal{T}\gets\mathcal{T}\cup\mathcal{C}$
    \EndWhile
    \State \Comment{tree construction complete}
    \State \Return $\textsc{TopB}(\mathcal{T},B)$
\EndProcedure
\end{algorithmic}
\end{algorithm}
\end{minipage}

\noindent\textbf{From the surrogate objective to priority-based expansion.}
Section~\ref{sec:preliminaries} shows that maximizing the surrogate 
objective in Eq.~\eqref{eq:surrogate-objective} reduces to selecting 
the top-$B$ prefixes by their log-probabilities under $\tilde{p}^\star$. Equivalently, the optimal 
tree consists of the $B$ prefixes $u$ with the largest 
$\log \tilde{p}^\star(u) = S(u)$, where $S$ is the path score in 
Eq.~\eqref{eq:path-score}. However, exhaustively expanding and ranking all candidate continuations is computationally infeasible due to the exponentially growing search 
space and large vocabulary branching factor. Instead, since the path 
score decomposes additively over depth, high-scoring partial prefixes 
are more likely to remain among the top-ranked candidates after further 
expansion. This naturally motivates a greedy priority-based expansion 
strategy for diffusion drafting: maintain a frontier of expandable nodes and iteratively expand 
the highest-scoring node.

\noindent We identify each tree node $v$ with the unique path from the root to 
$v$, denoted $P(v) = (t_1, \ldots, t_{d(v)})$, where 
$d(v) := |P(v)|$ is the depth of $v$. The cumulative path score is
$S(v) := S(P(v)) = \sum_{i=1}^{d(v)} s_{i, t_i}(c_i)$ with 
$c_i = (t_1, \ldots, t_{i-1})$. The root has 
$d(\text{root}) = 0, S(\text{root}) = 0$. At each iteration, the algorithm:
\begin{enumerate}
    \item Selects a frontier node $v^\star$ according to a search 
    policy $\pi$;
    \item Expands $v^\star$ by adding its top-$b$ children, ranked by 
    the token score $s_{d(v^\star)+1, t}(c_{d(v^\star)+1})$ at depth 
    $d(v^\star)+1$;
    \item Updates the frontier with the newly added children, removing 
    $v^\star$ from the expandable set.
\end{enumerate}
The iteration terminates when $|\mathcal{T}| = B$. Because $S$ is 
additive over depth, each child's cumulative score is computed 
incrementally from its parent's, avoiding redundant computation. 

\noindent We consider two widely used instantiations of priority-based
expansion under the path score $S$: beam search with global
retention (Algorithm~\ref{alg:beam}) and best-first search (BFS)
(Algorithm~\ref{alg:bfs}). Both prioritize high-$S$ partial paths
but differ in how the expansion budget is allocated across depth:
BFS maintains a global priority over the entire frontier, while
beam search restricts expansion to the top-$W$ in-beam nodes per
depth and retains out-of-beam nodes for the final top-$B$ selection.
Beam search can thus be interpreted as a width-constrained approximation to BFS.

\begin{wraptable}{r}{0.45\textwidth}
\vspace{-2em}
\centering
\scriptsize
\caption{Acceptance length comparison.}
\label{tab:search_comparison}
\begin{tabular}{lcccc}
\toprule
& \multicolumn{2}{c}{Qwen3-4B-DFlash}
& \multicolumn{2}{c}{Qwen3-8B-DFlash} \\
\cmidrule(lr){2-3} \cmidrule(lr){4-5}
Budget & BFS & Beam & BFS & Beam \\
\midrule
$B = 128$  & 9.21 & 9.27 & 9.25 & 9.28 \\
$B = 256$  & 9.55 & 9.54 & 9.60 & 9.62 \\
$B = 512$  & 9.81 & 9.79 & 9.90 & 9.88 \\
$B = 1024$ & 9.98 &  9.85 & 10.10 & 9.93 \\
\bottomrule
\end{tabular}
\end{wraptable}

\paragraph{BFS vs. Beam Search.}
Table~\ref{tab:search_comparison} reports acceptance length under
the two policies across budget sizes. At small to moderate budgets
($B \leq 512$), best-first and beam search yield similar acceptance length, as beam search's per-depth budget is sufficient to cover most high-priority expansions that best-first search would select under global prioritization. At $B = 1024$, best-first outperforms beam search by a small margin: with a larger budget, the preferred expansion pattern can become less uniform across depth, which a global priority handles more flexibly than a fixed per-depth beam. In practice, verification shifts from memory-bound to compute-bound
once $B$ exceeds a hardware-dependent threshold, after which
forward-pass cost grows linearly with $B$ and erodes the speedup.
We therefore operate at $B \leq 512$, which sits within the memory-bound regime where the two policies are comparable, and adopt beam search as the default search strategy.

\begin{figure}[t]
    \vspace{-0.5em}
    \centering
    \includegraphics[width=\linewidth]{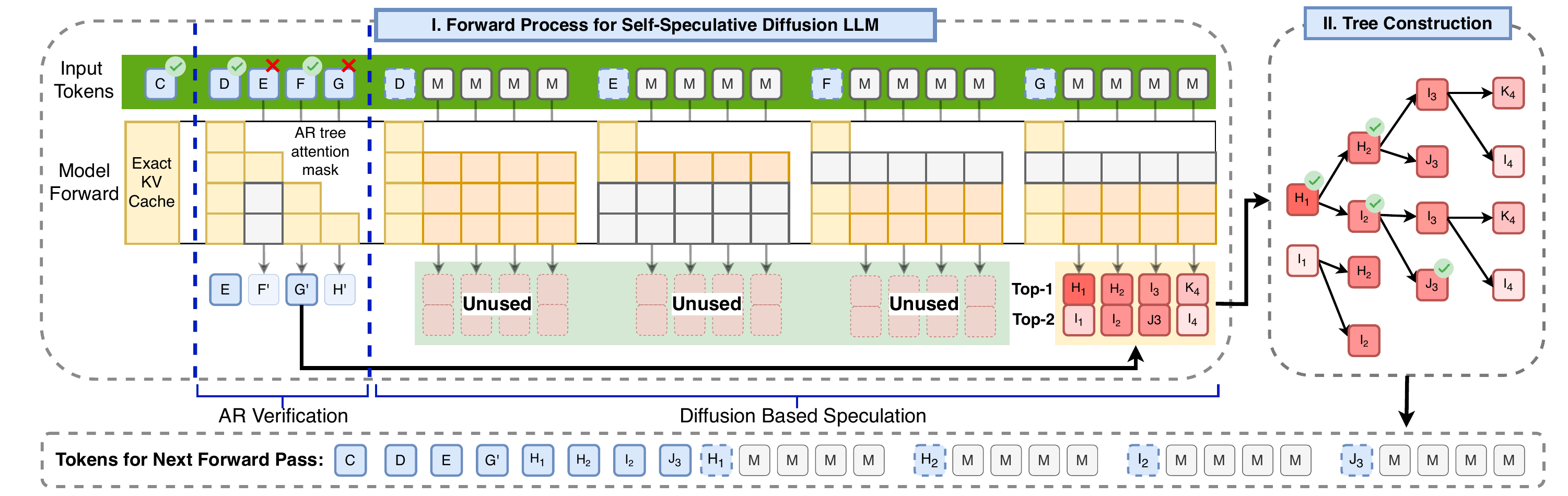}
    \vspace{-1.5em}
    \caption{Application of PRESTO to hybrid dLLMs under self-speculative decoding with a tree budget of 4 drafting tokens. The committed prefix reuses the exact KV cache for AR verification of the previous draft tree, while trailing mask-token slots are simultaneously denoised for diffusion-based speculation within the same forward pass. Speculative outputs attached to rejected branches are discarded, while valid speculation slots along the accepted path are used for next-round tree construction. The resulting draft tree is flattened into draft tokens for the next forward pass under AR tree attention. PRESTO is the first framework to explore tree-based drafting for self-speculative dLLMs. The detailed process is provided in Appendix~\ref{app:self_speculative}. }
    \label{fig:pipeline_speculative_dllm}
    \vspace{-0.5em}
\end{figure}

\subsection{Extending Tree Construction to hybrid dLLMs} 
\label{sec.self_speculative}
The proposed tree-construction framework can be naturally extended to support quadratic self-speculative decoding in hybrid dLLMs~\cite{liu2025tidarthinkdiffusiontalk,fu2026nemotron}\footnote{Nemotron-Labs-Diffusion primarily highlights a linear self-speculation decoding mode, where the model first generates draft tokens through diffusion decoding and then verifies them autoregressively using the same model. Applying PRESTO to this setting is straightforward and identical to the dedicated diffusion drafter setting discussed earlier, as PRESTO only modifies the diffusion drafting process while leaving the verification procedure unchanged. In addition to linear self-speculation, the Nemotron-Labs-Diffusion paper also introduces a quadratic self-speculation variant~\citep{fu2026nemotron} which is further explored in this work.}, as illustrated in Figure~\ref{fig:pipeline_speculative_dllm}. Quadratic self-speculation makes autoregressive verification and diffusion-based drafting parallelized within a single model forward pass: previously drafted tokens are verified autoregressively, while new draft candidates are simultaneously generated at trailing mask positions through diffusion decoding. 

\noindent PRESTO preserves this original verify-and-draft pipeline without introducing additional model forwards. During each forward pass, the committed prefix reuses the exact KV cache for AR verification of the previous draft tree, while the trailing mask-token slots attached to speculative branches are simultaneously denoised for next-round drafting. After verification, only the mask slots attached to the longest accepted path remain valid speculation positions, while the slots attached to rejected branches are discarded since their predictions were conditioned on prefixes that are no longer part of the committed context. PRESTO then extracts top-$k$ diffusion candidates from the valid mask slots and organizes them into a new draft tree using the proposed prefix-aligned scoring and priority-based tree expansion. The resulting tree is flattened into draft tokens and verified in the next forward pass using the AR tree attention mechanism.

\noindent This extension is particularly well-suited for self-speculative dLLMs because in standard linear self-speculative decoding, speculative tokens attached to rejected branches are discarded after verification, leading to substantial wasted computation. PRESTO alleviates this inefficiency by organizing diffusion outputs into a tree structure, allowing diverse speculative candidates generated within the same forward pass to be more effectively utilized during verification. As shown in Section~\ref{exp:results}, replacing linear drafting with PRESTO consistently improves acceptance length and decoding throughput across different datasets and model sizes.

\section{Empirical Validation}
In this section, we demonstrate that \name significantly improves the efficiency of diffusion-based speculative decoding across both standard and self-speculative diffusion based SD.

\subsection{Experiment Setup}
\textbf{Models and Tasks.} We evaluate \name on two settings:
(i) standard speculative decoding with a separate diffusion drafter
and AR target: dFlash with Qwen3-4B, Qwen3-8B, and Qwen3-Coder-30B-A3B
backbones; and (ii) hybrid dLLM: Nemotron-Labs-Diffusion-8B at both linear self-speculation and quadratic self-speculation modes. Evaluation spans three task categories:
\textbf{mathematical reasoning} (GSM8K~\cite{cobbe2021trainingverifierssolvemath},
MATH~\cite{lightman2023letsverifystepstep}, AIME24~\cite{aops2024aime},
AIME25~\cite{MAA_AIME_nd_a}), \textbf{code} (HumanEval~\cite{chen2021evaluatinglargelanguagemodels},
MBPP~\cite{austin2021programsynthesislargelanguage}, LiveCodeBench~\cite{jain2024livecodebenchholisticcontaminationfree}),
and \textbf{conversation} (MT-Bench~\cite{zheng2023judgingllmasajudgemtbenchchatbot},
Alpaca~\cite{alpaca}). We report average acceptance length ($\tau$) and
throughput (token/s) speedup over the AR baseline. All experiments
run on NVIDIA B200 GPUs.

\noindent \textbf{Implementations.} We implement PRESTO using two runtime backends. For dFlash and Nemotron-Labs-Diffusion under quadratic self-speculation, we use a PyTorch implementation with FlexAttention~\citep{dong2024flexattentionprogrammingmodel} for target model verification with tree-structured attention mask~\citep{Miao_2024,cai2024medusasimplellminference,li2024eagle2fasterinferencelanguage}. For Nemotron-Labs-Diffusion under linear self-speculation, we integrate PRESTO into the official SGLang implementation with FlashInfer attention~\citep{ye2025flashinferefficientcustomizableattention}. 
We instantiate the prefix-aligned score using a lightweight n-gram model with $n=3$\footnote{Following the n-gram model construction in~\citep{liu2026dartdiffusioninspiredspeculativedecoding}.}. Unless otherwise specified, we set $\lambda_d = 0.2$, batch size to 1, and use beam search with global retention, using beam width 10 and top-$k$ expansion ($k=10$) per position. We evaluate \name under both deterministic ($T=0$) and stochastic ($T=1$) decoding. For dFlash, we sweep tree budgets $\mathcal{T} \in \{128,256,512\}$ and report the best-performing configuration in terms of end-to-end throughput. For Nemotron-Labs-Diffusion, we use a fixed tree budget $\mathcal{T}$ of 32 following its speculative block size.

\subsection{Experiment Results}
\label{exp:results}
\begin{table*}[t]
\centering
\small
\setlength{\tabcolsep}{2pt}
\caption{Decoding speedup over baseline and average acceptance length ($\tau$) on Qwen3 models with thinking mode disabled and a maximum of 2048 generated tokens.}
\label{tab:main-results}
\resizebox{\linewidth}{!}{%
\begin{tabular}{c l @{\hspace{0.3em}} cc cc cc cc @{\hspace{0.3em}} cc cc cc cc @{\hspace{0.3em}} cc cc @{\hspace{0.3em}} cc}
\toprule
\multirow{2}{*}{Model} & \multirow{2}{*}{Method}
& \multicolumn{8}{c@{\hspace{0.3em}}}{\textsc{Math}}
& \multicolumn{8}{c@{\hspace{0.3em}}}{\textsc{Code}}
& \multicolumn{4}{c@{\hspace{0.3em}}}{\textsc{Chat}}
& \multicolumn{2}{c}{\textit{Avg.}} \\
\cmidrule(lr){3-10} \cmidrule(lr){11-18} \cmidrule(lr){19-22} \cmidrule(lr){23-24}
& & \multicolumn{2}{c}{GSM8K}
& \multicolumn{2}{c}{MATH-500}
& \multicolumn{2}{c}{AIME24}
& \multicolumn{2}{c@{\hspace{1.2em}}}{AIME25}
& \multicolumn{2}{c}{HumanEval}
& \multicolumn{2}{c}{MBPP}
& \multicolumn{2}{c}{LCB}
& \multicolumn{2}{c@{\hspace{1.2em}}}{SWE-Bench}
& \multicolumn{2}{c}{MT-Bench}
& \multicolumn{2}{c@{\hspace{1.2em}}}{Alpaca}
& \multicolumn{2}{c}{\textit{Avg.}} \\
\midrule

\multicolumn{2}{c}{Temperature = 0}
& {\scriptsize Sp.} & {\scriptsize $\tau$}
& {\scriptsize Sp.} & {\scriptsize $\tau$}
& {\scriptsize Sp.} & {\scriptsize $\tau$}
& {\scriptsize Sp.} & {\scriptsize $\tau$}
& {\scriptsize Sp.} & {\scriptsize $\tau$}
& {\scriptsize Sp.} & {\scriptsize $\tau$}
& {\scriptsize Sp.} & {\scriptsize $\tau$}
& {\scriptsize Sp.} & {\scriptsize $\tau$}
& {\scriptsize Sp.} & {\scriptsize $\tau$}
& {\scriptsize Sp.} & {\scriptsize $\tau$}
& {\scriptsize Sp.} & {\scriptsize $\tau$} \\
\midrule

\multirow{3}{*}{4B}
& dFlash
& 5.1$\times$ & 6.5 & 6.5$\times$ & 7.9 & 6.5$\times$ & 7.4 & 6.2$\times$ & 7.4
& 5.6$\times$ & 6.7 & 5.0$\times$ & 6.1 & 5.9$\times$ & 6.9 & 3.2$\times$ & 3.6
& 2.9$\times$ & 4.4 & 2.4$\times$ & 3.1
& 4.9$\times$ & 6.0 \\
& \name
& 7.9$\times$ & 9.4 & 9.2$\times$ & 10.9 & 8.7$\times$ & 10.5 & 8.8$\times$ & 10.3
& 8.1$\times$ & 9.9 & 7.9$\times$ & 9.3 & 8.4$\times$ & 10.0 & 5.3$\times$ & 5.9
& 4.9$\times$ & 6.7 & 3.9$\times$ & 5.0
& 7.3$\times$ & 8.8 \\
\cmidrule(lr){2-24}
& \textbf{$\times$ Gain}
% \rowcolor[rgb]{0.88,0.96,0.84}
& \textbf{1.55$\times$} & 
& \textbf{1.42$\times$} & 
& \textbf{1.34$\times$} & 
& \textbf{1.42$\times$} & 
& \textbf{1.45$\times$} & 
& \textbf{1.58$\times$} & 
& \textbf{1.42$\times$} & 
& \textbf{1.66$\times$} & 
& \textbf{1.69$\times$} & 
& \textbf{1.62$\times$} & 
& \textbf{1.48$\times$} & \\

\midrule
\multirow{3}{*}{8B}
& dFlash
& 5.0$\times$ & 6.6 & 6.3$\times$ & 7.9 & 6.2$\times$ & 7.5 & 6.0$\times$ & 7.0
& 5.4$\times$ & 6.5 & 4.8$\times$ & 6.0 & 5.7$\times$ & 7.1 & 3.3$\times$ & 3.6
& 2.8$\times$ & 4.3 & 2.4$\times$ & 3.1
& 4.8$\times$ & 6.0 \\
& \name
& 8.0$\times$ & 9.6 & 9.4$\times$ & 11.1 & 8.5$\times$ & 10.7 & 8.7$\times$ & 10.2
& 8.5$\times$ & 9.9 & 7.7$\times$ & 9.4 & 8.5$\times$ & 10.3 & 5.0$\times$ & 6.0
& 4.8$\times$ & 6.6 & 3.9$\times$ & 5.0
& 7.3$\times$ & 8.9 \\
\cmidrule(lr){2-24}
& \textbf{$\times$ Gain}
% \rowcolor[rgb]{0.88,0.96,0.84}
& \textbf{1.60$\times$} & 
& \textbf{1.49$\times$} &
& \textbf{1.37$\times$} & 
& \textbf{1.45$\times$} & 
& \textbf{1.57$\times$} &
& \textbf{1.60$\times$} & 
& \textbf{1.49$\times$} & 
& \textbf{1.52$\times$} &
& \textbf{1.71$\times$} & 
& \textbf{1.62$\times$} & 
& \textbf{1.52$\times$} &\\

\midrule
\multirow{3}{*}{30B}
& dFlash
& 3.1$\times$ & 5.2 & 4.1$\times$ & 5.6 & 2.8$\times$ & 5.3 & 2.7$\times$ & 5.1
& 5.6$\times$ & 8.0 & 5.6$\times$ & 7.2 & 3.6$\times$ & 6.2 & 3.2$\times$ & 3.6
& 2.2$\times$ & 3.5 & 1.8$\times$ & 2.2
& 3.5$\times$ & 5.2 \\
& \name
& 4.1$\times$ & 7.9 & 5.4$\times$ & 8.0 & 3.6$\times$ & 8.0 & 3.6$\times$ & 7.9
& 7.0$\times$ & 10.9 & 7.5$\times$ & 9.9 & 4.7$\times$ & 8.5 & 4.6$\times$ & 5.6
& 3.0$\times$ & 5.2 & 2.5$\times$ & 3.4
& 4.6$\times$ & 7.5 \\
\cmidrule(lr){2-24}
& \textbf{$\times$ Gain}
% \rowcolor[rgb]{0.88,0.96,0.84}
& \textbf{1.32$\times$} & 
& \textbf{1.32$\times$} & 
& \textbf{1.29$\times$} & 
& \textbf{1.33$\times$} &
& \textbf{1.25$\times$} &
& \textbf{1.34$\times$} &
& \textbf{1.31$\times$} &
& \textbf{1.44$\times$} & 
& \textbf{1.36$\times$} &
& \textbf{1.39$\times$} & 
& \textbf{1.33$\times$} &\\

\midrule
\multicolumn{2}{c}{Temperature = 1}
& {\scriptsize Sp.} & {\scriptsize $\tau$}
& {\scriptsize Sp.} & {\scriptsize $\tau$}
& {\scriptsize Sp.} & {\scriptsize $\tau$}
& {\scriptsize Sp.} & {\scriptsize $\tau$}
& {\scriptsize Sp.} & {\scriptsize $\tau$}
& {\scriptsize Sp.} & {\scriptsize $\tau$}
& {\scriptsize Sp.} & {\scriptsize $\tau$}
& {\scriptsize Sp.} & {\scriptsize $\tau$}
& {\scriptsize Sp.} & {\scriptsize $\tau$}
& {\scriptsize Sp.} & {\scriptsize $\tau$}
& {\scriptsize Sp.} & {\scriptsize $\tau$} \\
\midrule

\multirow{3}{*}{4B}
& dFlash
& 4.7$\times$ & 6.0 & 5.2$\times$ & 6.6 & 3.8$\times$ & 5.0 & 3.9$\times$ & 4.9
& 4.8$\times$ & 6.0 & 4.4$\times$ & 5.6 & 5.0$\times$ & 6.6 & 2.5$\times$ & 3.1
& 2.7$\times$ & 4.1 & 2.2$\times$ & 3.0
& 3.9$\times$ & 5.1 \\
& \name
& 7.2$\times$ & 9.1 & 7.6$\times$ & 9.8 & 6.0$\times$ & 7.7 & 6.1$\times$ & 7.7
& 7.6$\times$ & 9.3 & 7.0$\times$ & 9.0 & 6.7$\times$ & 9.4 & 4.2$\times$ & 5.0
& 4.3$\times$ & 6.2 & 3.7$\times$ & 4.7
& 6.0$\times$ & 7.8 \\
\cmidrule(lr){2-24}
& \textbf{$\times$ Gain}
% \rowcolor[rgb]{0.88,0.96,0.84}
& \textbf{1.53$\times$} & 
& \textbf{1.46$\times$} &
& \textbf{1.58$\times$} &
& \textbf{1.56$\times$} & 
& \textbf{1.58$\times$} &
& \textbf{1.59$\times$} & 
& \textbf{1.34$\times$} &
& \textbf{1.68$\times$} &
& \textbf{1.59$\times$} &
& \textbf{1.68$\times$} & 
& \textbf{1.54$\times$} & \\

\midrule
\multirow{3}{*}{8B}
& dFlash
& 4.8$\times$ & 6.0 & 5.0$\times$ & 6.6 & 3.9$\times$ & 5.1 & 3.8$\times$ & 5.0
& 4.4$\times$ & 5.4 & 4.1$\times$ & 5.2 & 5.2$\times$ & 6.8 & 2.3$\times$ & 2.8
& 2.6$\times$ & 3.8 & 2.1$\times$ & 2.9
& 3.8$\times$ & 5.0 \\
& \name
& 7.4$\times$ & 9.0 & 7.4$\times$ & 9.6 & 6.3$\times$ & 7.8 & 6.5$\times$ & 7.7
& 7.2$\times$ & 8.6 & 6.9$\times$ & 8.6 & 7.0$\times$ & 9.8 & 4.0$\times$ & 4.7
& 4.2$\times$ & 5.9 & 3.6$\times$ & 4.7
& 6.0$\times$ & 7.6 \\
\cmidrule(lr){2-24}
& \textbf{$\times$ Gain}
% \rowcolor[rgb]{0.88,0.96,0.84}
& \textbf{1.54$\times$} &
& \textbf{1.48$\times$} &
& \textbf{1.62$\times$} &
& \textbf{1.71$\times$} &
& \textbf{1.64$\times$} &
& \textbf{1.68$\times$} &
& \textbf{1.35$\times$} &
& \textbf{1.74$\times$} &
& \textbf{1.62$\times$} & 
& \textbf{1.71$\times$} & 
& \textbf{1.58$\times$} & \\

\midrule
\multirow{3}{*}{30B}
& dFlash
& 4.1$\times$ & 5.1 & 4.2$\times$ & 5.3 & 3.3$\times$ & 4.3 & 3.3$\times$ & 4.2
& 5.8$\times$ & 7.7 & 5.6$\times$ & 7.1 & 3.8$\times$ & 5.6 & 2.7$\times$ & 3.2
& 2.1$\times$ & 3.2 & 1.8$\times$ & 2.1
& 3.7$\times$ & 4.8 \\
& \name
& 4.5$\times$ & 7.8 & 5.5$\times$ & 7.7 & 3.7$\times$ & 6.7 & 3.4$\times$ & 6.5
& 6.8$\times$ & 10.2 & 7.9$\times$ & 9.6 & 4.3$\times$ & 8.0 & 4.4$\times$ & 4.9
& 3.0$\times$ & 5.0 & 2.6$\times$ & 3.3
& 4.6$\times$ & 7.0 \\
\cmidrule(lr){2-24}
& \textbf{$\times$ Gain}
% \rowcolor[rgb]{0.88,0.96,0.84}
& \textbf{1.10$\times$} & 
& \textbf{1.31$\times$} & 
& \textbf{1.12$\times$} &
& \textbf{1.03$\times$} & 
& \textbf{1.17$\times$} &
& \textbf{1.41$\times$} &
& \textbf{1.13$\times$} & 
& \textbf{1.63$\times$} &
& \textbf{1.43$\times$} &
& \textbf{1.44$\times$} &
& \textbf{1.26$\times$} & \\

\bottomrule
\end{tabular}%
}
\end{table*}

\textbf{\name on dFlash.} As illustrated in Table~\ref{tab:main-results}, \name consistently achieves substantial improvements in the average acceptance length $\tau$ across all models and tasks, which directly translates into higher end-to-end throughput speedup. For example, on Qwen3-4B with $T=0$, $\tau$ increases from $7.9$ to $10.9$ on MATH-500 and from $7.4$ to $10.3$ on AIME25, while on code benchmarks it improves from $6.9$ to $10.0$ on LCB and from $6.7$ to $9.9$ on HumanEval, leading to corresponding speedup gains across all settings. The largest improvements are observed on math and code benchmarks, where $\tau$ typically increases by $+2$ to $+3$ tokens. These trends remain stable across model scales and decoding regimes: even for stronger drafters such as Qwen3-8B and under stochastic decoding ($T=1$), \name continues to deliver comparable relative gains. These results indicate that \name effectively pushes the acceptance length of dFlash closer to its upper bound, delivering consistent and substantial end-to-end throughput speedup gains. The end-to-end throughput results based on torch implementation is shown in Appendix~\ref{app:throughput_dflash}. 

\begin{table*}[hth]
\centering
\small
\setlength{\tabcolsep}{2pt}
\caption{Decoding speedup over baseline, average acceptance length ($\tau$), and absolute throughput on Nemotron-Labs-Diffusion-8B at linear self speculation decoding mode with a maximum of 2048 generated tokens.}
\label{tab:main-results-linear-ss}
\resizebox{\linewidth}{!}{%
\begin{tabular}{
c l
@{\hspace{0.3em}} cc cc cc cc
@{\hspace{0.3em}} cc cc cc cc
@{\hspace{0.3em}} cc cc
@{\hspace{0.3em}} ccc
}
\toprule
\multirow{2}{*}{Model} & \multirow{2}{*}{Method}
& \multicolumn{8}{c@{\hspace{1.2em}}}{\sc{Math}}
& \multicolumn{8}{c@{\hspace{1.2em}}}{\sc{Code}}
& \multicolumn{4}{c@{\hspace{1.2em}}}{\sc{Chat}}
& \multicolumn{3}{c}{\textit{Avg.}} \\
\cmidrule(lr){3-10} \cmidrule(lr){11-18} \cmidrule(lr){19-22} \cmidrule(lr){23-25}

& & \multicolumn{2}{c}{GSM8K}
& \multicolumn{2}{c}{MATH-500}
& \multicolumn{2}{c}{AIME24}
& \multicolumn{2}{c@{\hspace{1.2em}}}{AIME25}
& \multicolumn{2}{c}{HumanEval}
& \multicolumn{2}{c}{MBPP}
& \multicolumn{2}{c}{LCB}
& \multicolumn{2}{c@{\hspace{1.2em}}}{SWE-Bench}
& \multicolumn{2}{c}{MT-Bench}
& \multicolumn{2}{c}{Alpaca}
& \multicolumn{3}{c}{\textit{Avg.}} \\
\midrule

\multicolumn{2}{c}{Temperature = 0}
& \tabsmall Sp. & \tabsmall $\tau$
& \tabsmall Sp. & \tabsmall $\tau$
& \tabsmall Sp. & \tabsmall $\tau$
& \tabsmall Sp. & \tabsmall $\tau$
& \tabsmall Sp. & \tabsmall $\tau$
& \tabsmall Sp. & \tabsmall $\tau$
& \tabsmall Sp. & \tabsmall $\tau$
& \tabsmall Sp. & \tabsmall $\tau$
& \tabsmall Sp. & \tabsmall $\tau$
& \tabsmall Sp. & \tabsmall $\tau$
& \tabsmall Sp. & \tabsmall $\tau$ & \tabsmall Thr. \\
\midrule

\multirow{3}{*}{8B}
& vanilla
& 5.7$\times$ & 10.8 & 6.4$\times$ & 12.2 & 5.7$\times$ & 11.0 & 5.5$\times$ & 10.5
& 5.0$\times$ & 9.3 & 4.4$\times$ & 8.2 & 4.7$\times$ & 9.0 & 3.7$\times$ & 6.8
& 2.8$\times$ & 5.2 & 2.5$\times$ & 4.6 & 4.6$\times$ & 8.8 & 642 \\

& \name
& 6.0$\times$ & 11.9 & 6.7$\times$ & 13.2 & 6.1$\times$ & 12.2 & 5.8$\times$ & 11.5
& 5.2$\times$ & 10.3 & 4.6$\times$ & 9.1 & 4.9$\times$ & 10.0 & 4.0$\times$ & 7.9
& 3.2$\times$ & 6.5 & 2.9$\times$ & 5.8 & 4.9$\times$ & 9.9 & 686 \\

\cmidrule(lr){2-25}
& \textbf{$\times$ Gain}
% \rowcolor[rgb]{0.88,0.96,0.84}
& \textbf{1.05$\times$} &
& \textbf{1.04$\times$} &
& \textbf{1.07$\times$} &
& \textbf{1.06$\times$} &
& \textbf{1.04$\times$} &
& \textbf{1.04$\times$} &
& \textbf{1.04$\times$} &
& \textbf{1.08$\times$} &
& \textbf{1.14$\times$} &
& \textbf{1.16$\times$} &
& \textbf{1.06$\times$} & & \\

\midrule

\multicolumn{2}{c}{Temperature = 1}
& \tabsmall Sp. & \tabsmall $\tau$
& \tabsmall Sp. & \tabsmall $\tau$
& \tabsmall Sp. & \tabsmall $\tau$
& \tabsmall Sp. & \tabsmall $\tau$
& \tabsmall Sp. & \tabsmall $\tau$
& \tabsmall Sp. & \tabsmall $\tau$
& \tabsmall Sp. & \tabsmall $\tau$
& \tabsmall Sp. & \tabsmall $\tau$
& \tabsmall Sp. & \tabsmall $\tau$
& \tabsmall Sp. & \tabsmall $\tau$
& \tabsmall Sp. & \tabsmall $\tau$ & \tabsmall Thr. \\

\midrule

\multirow{3}{*}{8B}
& vanilla
& 2.9$\times$ & 5.6 & 2.9$\times$ & 5.7 & 2.5$\times$ & 5.0 & 2.6$\times$ & 4.9
& 2.3$\times$ & 4.6 & 2.1$\times$ & 4.2 & 2.0$\times$ & 4.0 & 1.4$\times$ & 2.9
& 1.4$\times$ & 2.7 & 1.2$\times$ & 2.3 & 2.1$\times$ & 4.2 & 290 \\

& \name
& 3.1$\times$ & 6.1 & 3.3$\times$ & 7.1 & 2.9$\times$ & 5.7 & 2.9$\times$ & 6.4
& 2.7$\times$ & 5.5 & 2.6$\times$ & 5.0 & 2.3$\times$ & 4.8 & 1.8$\times$ & 3.6
& 1.8$\times$ & 3.5 & 1.5$\times$ & 3.2 & 2.5$\times$ & 5.1 & 339 \\

\cmidrule(lr){2-25}
& \textbf{$\times$ Gain}
% \rowcolor[rgb]{0.88,0.96,0.84}
& \textbf{1.07$\times$} &
& \textbf{1.14$\times$} &
& \textbf{1.16$\times$} &
& \textbf{1.12$\times$} &
& \textbf{1.17$\times$} &
& \textbf{1.24$\times$} &
& \textbf{1.15$\times$} &
& \textbf{1.29$\times$} &
& \textbf{1.29$\times$} &
& \textbf{1.25$\times$} &
& \textbf{1.17$\times$} & & \\

\bottomrule
\end{tabular}%
}
\end{table*}

\begin{table*}[t]
\centering
\small
\setlength{\tabcolsep}{2pt}
\caption{Decoding speedup over baseline, average acceptance length ($\tau$), and absolute throughput on Nemotron-Labs-Diffusion-8B at quadratic self speculation decoding mode with a maximum of 2048 generated tokens.}
\label{tab:main-results-small}
\resizebox{\linewidth}{!}{%
\begin{tabular}{
c l
@{\hspace{0.3em}} cc cc cc cc
@{\hspace{0.3em}} cc cc cc cc
@{\hspace{0.3em}} cc cc
@{\hspace{0.3em}} ccc
}
\toprule
\multirow{2}{*}{Model} & \multirow{2}{*}{Method}
& \multicolumn{8}{c@{\hspace{1.2em}}}{\sc{Math}}
& \multicolumn{8}{c@{\hspace{1.2em}}}{\sc{Code}}
& \multicolumn{4}{c@{\hspace{1.2em}}}{\sc{Chat}}
& \multicolumn{3}{c}{\textit{Avg.}} \\
\cmidrule(lr){3-10}
\cmidrule(lr){11-18}
\cmidrule(lr){19-22}
\cmidrule(lr){23-25}

& & \multicolumn{2}{c}{GSM8K}
& \multicolumn{2}{c}{MATH-500}
& \multicolumn{2}{c}{AIME24}
& \multicolumn{2}{c@{\hspace{1.2em}}}{AIME25}
& \multicolumn{2}{c}{HumanEval}
& \multicolumn{2}{c}{MBPP}
& \multicolumn{2}{c}{LCB}
& \multicolumn{2}{c@{\hspace{1.2em}}}{SWE-Bench}
& \multicolumn{2}{c}{MT-Bench}
& \multicolumn{2}{c}{Alpaca}
& \multicolumn{3}{c}{\textit{Avg.}} \\
\midrule

\multicolumn{2}{c}{Temperature = 0}
& \tabsmall Sp. & \tabsmall $\tau$
& \tabsmall Sp. & \tabsmall $\tau$
& \tabsmall Sp. & \tabsmall $\tau$
& \tabsmall Sp. & \tabsmall $\tau$
& \tabsmall Sp. & \tabsmall $\tau$
& \tabsmall Sp. & \tabsmall $\tau$
& \tabsmall Sp. & \tabsmall $\tau$
& \tabsmall Sp. & \tabsmall $\tau$
& \tabsmall Sp. & \tabsmall $\tau$
& \tabsmall Sp. & \tabsmall $\tau$
& \tabsmall Sp. & \tabsmall $\tau$
& \tabsmall Thr. \\
\midrule

\multirow{3}{*}{8B}
& vanilla
& 2.6$\times$ & 4.7 & 2.7$\times$ & 6.8 & 3.2$\times$ & 5.3 & 3.3$\times$ & 5.7
& 3.2$\times$ & 4.8 & 3.1$\times$ & 4.5 & 2.4$\times$ & 4.7 & 1.8$\times$ & 3.5
& 2.1$\times$ & 3.0 & 1.8$\times$ & 2.7 & 2.6$\times$ & 4.7
& 94 \\

& \name
& 5.3$\times$ & 6.5 & 4.4$\times$ & 7.5 & 3.5$\times$ & 6.1 & 3.7$\times$ & 6.4
& 3.7$\times$ & 5.8 & 3.7$\times$ & 5.4 & 3.1$\times$ & 5.6 & 2.7$\times$ & 4.7
& 2.8$\times$ & 4.2 & 2.3$\times$ & 3.6 & 3.5$\times$ & 5.6
& 127 \\

\cmidrule(lr){2-25}
& \textbf{$\times$ Gain}
% \rowcolor[rgb]{0.88,0.96,0.84}
& \textbf{2.04$\times$} &
& \textbf{1.63$\times$} &
& \textbf{1.09$\times$} &
& \textbf{1.12$\times$} &
& \textbf{1.16$\times$} &
& \textbf{1.19$\times$} &
& \textbf{1.29$\times$} &
& \textbf{1.50$\times$} &
& \textbf{1.33$\times$} &
& \textbf{1.28$\times$} &
& \textbf{1.35$\times$} &
& \\

\midrule

\multicolumn{2}{c}{Temperature = 1}
& \tabsmall Sp. & \tabsmall $\tau$
& \tabsmall Sp. & \tabsmall $\tau$
& \tabsmall Sp. & \tabsmall $\tau$
& \tabsmall Sp. & \tabsmall $\tau$
& \tabsmall Sp. & \tabsmall $\tau$
& \tabsmall Sp. & \tabsmall $\tau$
& \tabsmall Sp. & \tabsmall $\tau$
& \tabsmall Sp. & \tabsmall $\tau$
& \tabsmall Sp. & \tabsmall $\tau$
& \tabsmall Sp. & \tabsmall $\tau$
& \tabsmall Sp. & \tabsmall $\tau$
& \tabsmall Thr. \\

\midrule

\multirow{3}{*}{8B}
& vanilla
& 1.4$\times$ & 3.0 & 1.8$\times$ & 3.4 & 1.5$\times$ & 2.6 & 1.7$\times$ & 3.0
& 1.7$\times$ & 2.6 & 1.6$\times$ & 2.4 & 1.2$\times$ & 2.3 & 1.0$\times$ & 1.8
& 1.2$\times$ & 1.7 & 1.0$\times$ & 1.5 & 1.4$\times$ & 2.4
& 51 \\

& \name
& 4.1$\times$ & 5.7 & 4.0$\times$ & 6.6 & 2.9$\times$ & 5.1 & 3.3$\times$ & 5.7
& 3.2$\times$ & 4.9 & 3.0$\times$ & 4.4 & 2.8$\times$ & 4.8 & 2.2$\times$ & 3.7
& 2.3$\times$ & 3.5 & 2.0$\times$ & 3.0 & 3.0$\times$ & 4.7
& 109 \\

\cmidrule(lr){2-25}
& \textbf{$\times$ Gain}
% \rowcolor[rgb]{0.88,0.96,0.84}
& \textbf{2.93$\times$} &
& \textbf{2.22$\times$} &
& \textbf{1.93$\times$} &
& \textbf{1.94$\times$} &
& \textbf{1.88$\times$} &
& \textbf{1.88$\times$} &
& \textbf{2.33$\times$} &
& \textbf{2.20$\times$} &
& \textbf{1.92$\times$} &
& \textbf{2.00$\times$} &
& \textbf{2.14$\times$} &
& \\

\bottomrule
\end{tabular}%
}
\end{table*}

\noindent \textbf{\name on Nemotron-Labs-Diffusion (NLD).} As shown in Table~\ref{tab:main-results-linear-ss}, \name consistently improves both throughput and average acceptance length over the official linear self-speculative decoding implementation of Nemotron-Labs-Diffusion across all benchmarks and decoding temperatures. By replacing linear drafting with tree-based exploration while preserving the original verification pipeline, \name remains effective across a wide range of tasks. Under deterministic decoding ($T=0$), \name increases average end-to-end throughput from $4.6\times$ to $4.9\times$ while improving average acceptance length from $8.8$ to $9.9$, with consistent gains observed on all benchmarks. Under stochastic decoding ($T=1$), where a single drafted trajectory is more likely to diverge from the target model, the benefits become more pronounced: average end-to-end throughput improves from $2.1\times$ to $2.5\times$, while average acceptance length increases from $4.2$ to $5.1$. Across individual benchmarks, \name achieves up to $1.29\times$ end-to-end throughput improvement and increases acceptance length by as much as $39\%$ on Alpaca under stochastic decoding (T=1). These results demonstrate that the benefits of tree-based drafting persist even within a production-grade, highly optimized inference engine. The end-to-end throughput results based on SGLang is shown in Appendix~\ref{app:throughput_nld}.

\noindent \textbf{\name on Nemotron-Labs-Diffusion (NLD) at quadratic self speculation mode.} We further evaluate \name under the quadratic self-speculative decoding setting introduced in Nemotron-Labs-Diffusion. As Table~\ref{tab:main-results-small} shows, across all benchmarks and decoding settings, \name also consistently outperforms vanilla quadratic self-speculative decoding in both end-to-end throughput and average acceptance length. 
Under deterministic decoding ($T=0$), \name provides stable gains across math, code, and chat tasks, improving average speedup from $2.6\times$ to $3.5\times$ on NLD-8B, while also consistently increasing acceptance length from $4.7$ to $5.6$. The largest gains are observed on math benchmarks, where \name achieves up to $2.04\times$ on GSM8k for NLD-8B. Gains are largest under stochastic decoding ($T=1$), where single-path drafting degrades sharply from trajectory mismatch and accumulated uncertainty, while \name's tree-structured exploration retains multiple plausible continuations. On NLD-8B, this raises average end-to-end speedup from $1.4\times$ to $3.0\times$ and nearly doubles acceptance length from $2.4$ to $4.7$, reaching up to $2.9\times$ throughput and over $2\times$ acceptance length over vanilla self-speculative decoding. These positive results indicate the significant benefits of applying PRESTO to quadratic self-speculation.

\noindent We note that the throughput for dFlash and NLD quadratic self speculation can be further improved through system-level optimizations, including custom kernels, optimized KV-cache management, and efficient request scheduling. The goal here is not to fully optimize absolute throughput, but rather to provide a first-hand comparison between linear and tree-based drafting for diffusion drafters under the same native PyTorch implementation.
\section{Discussions and Ablations}
\begin{figure}[t]
    \centering

    \includegraphics[width=\linewidth]{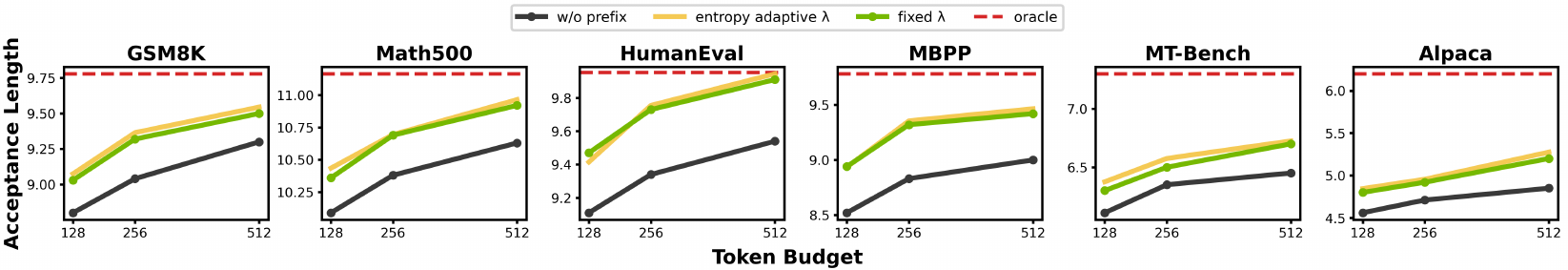}
    \vspace{-1em}

    \includegraphics[width=\linewidth]{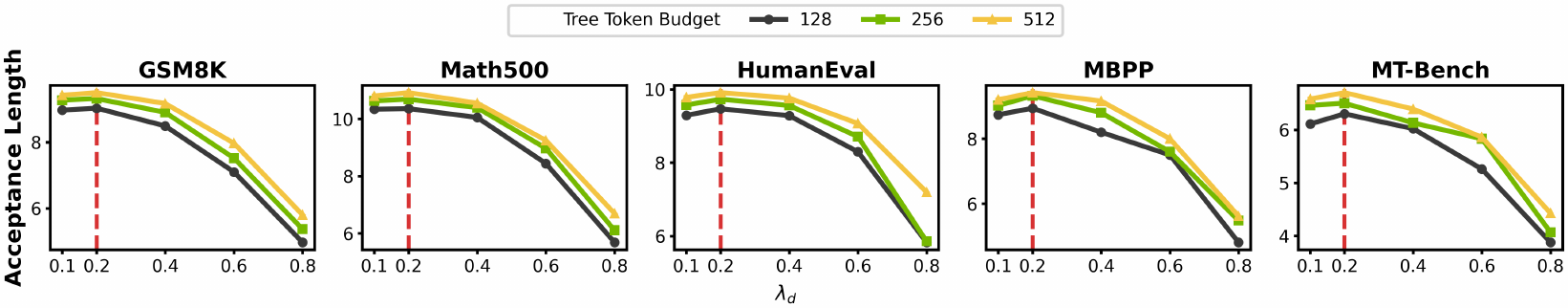}
    \vspace{-1em}

    \includegraphics[width=\linewidth]{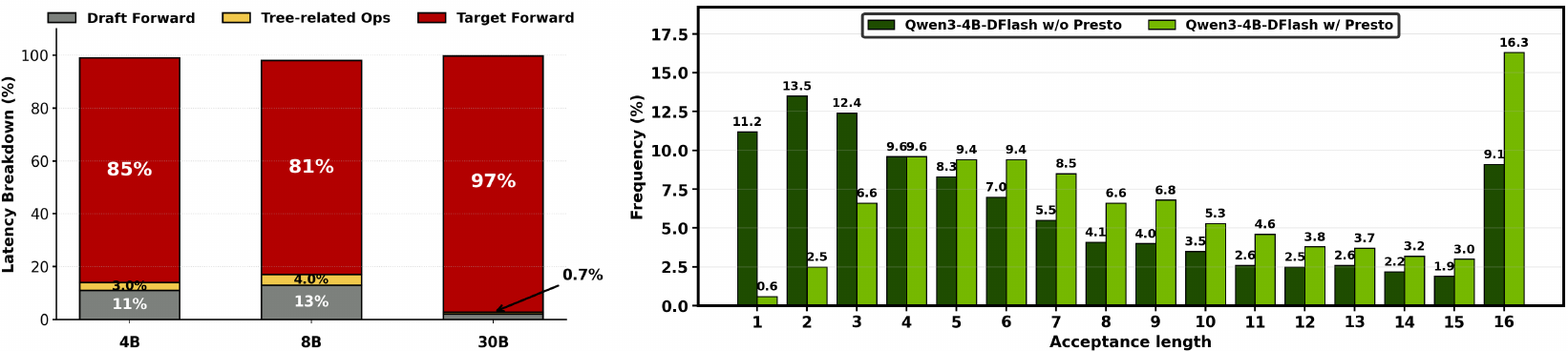}

    % \vspace{-1em}
\caption{
\textbf{Top}: Effectiveness of prefix-conditioned compatibility signals. Incorporating the prefix-conditioned term $\rho_d$ consistently improves acceptance length across datasets and tree token budgets.
\textbf{Middle}: Sensitivity analysis of $\lambda_d$ under different datasets and tree token budgets. Moderate values of $\lambda_d$ consistently achieve the best performance, with $\lambda_d \approx 0.2$ providing the most stable gains across settings.
\textbf{Bottom left}: Per-step latency breakdown across different model scales. Target-model verification dominates the overall decoding cost, while tree-related operations introduce only negligible overhead.
\textbf{Bottom right}: Distribution of acceptance lengths on GSM8K. \name significantly shifts the distribution toward longer accepted prefixes and substantially increases the frequency of near-complete and full-block acceptance. All results are obtained using Qwen3-4B-DFlash with diffusion block size 16.
}
    \label{fig:ablation_all}
    % \vspace{-0.5em}
\end{figure}

\textbf{Effectiveness of prefix-conditioned signal.}
We first ablate the contribution of the prefix-conditioned compatibility term $\rho_d$ by comparing our full scoring function against a variant that only relies on the diffusion marginal confidence $q_d$. As shown in Figure~\ref{fig:ablation_all} top, incorporating $\rho_d$ consistently improves the average acceptance length across all tasks and tree token budgets. The gains are particularly noticeable on reasoning and coding benchmarks such as GSM8K, Math500, and HumanEval, where prefix consistency plays a more important role in determining downstream verification success. These results suggest that diffusion marginal confidence alone is insufficient for reliably ranking candidate paths, while the prefix-conditioned signal provides complementary information that better aligns tree expansion with the autoregressive verification process.

\noindent\textbf{Sensitivity of $\lambda_{d}$.} Figure~\ref{fig:ablation_all} middle shows the impact of varying $\lambda_d$ under different datasets and tree token budgets. Across nearly all settings, moderate values of $\lambda_d$ consistently achieve the best acceptance length, with $\lambda_d \approx 0.2$ providing the most stable performance. When $\lambda_d$ is too small, the prefix-conditioned signal is underutilized and the scoring function degenerates toward pure diffusion confidence ranking. In contrast, overly large $\lambda_d$ overemphasizes $\rho_d$ and suppresses the diffusion marginal confidence $q_d$, leading to unstable candidate ranking and substantial degradation in acceptance length. Importantly, the optimal range remains highly consistent across different tasks and tree sizes, suggesting that the proposed scoring function is robust.

\noindent\textbf{Entropy Adaptive $\lambda_{d}$ vs. Fixed $\lambda_{d}$.}
We further explore an entropy-adaptive $\lambda_{d}$ that dynamically balances $q_d$ and $\rho_d$ according to the uncertainty of the diffusion drafter: confident predictions rely more on $q_d$, while uncertain predictions assign larger weights to $\rho_d$ (details in Appendix~\ref{app:entropy-adaptive}). As shown in Figure~\ref{fig:ablation_all} top, entropy-adaptive $\lambda_d$ provides only marginal improvements over a fixed $\lambda_d$, and the overall trends remain highly similar across datasets. This suggests that although uncertainty-aware weighting is intuitively appealing, a simple fixed $\lambda_d$ is already sufficient to capture most of the benefits of prefix-conditioned ranking in practice, while avoiding additional complexity.

\noindent\textbf{Overhead Analysis.} As shown in Figure~\ref{fig:ablation_all} bottom left, target-model verification overwhelmingly dominates the overall decoding latency across all model scales, accounting for 81\%--97\% of the total runtime. In comparison, tree-related operations, including tree construction, candidate filtering, and attention-mask preparation, contribute only 0.7\%--4\% of the per-step latency. This demonstrates that the proposed tree construction introduces negligible system overhead relative to the target-model forward pass. Moreover, the overhead becomes proportionally smaller for larger models, indicating favorable scalability of the framework.

\noindent\textbf{Acceptance Length Distribution Analysis.} Figure~\ref{fig:ablation_all} bottom right further illustrates how \name reshapes the acceptance-length distribution. Compared with vanilla diffusion drafting, \name substantially reduces short-acceptance cases while significantly increasing the frequency of long-prefix and full-block acceptance. In particular, the probability of achieving near-complete or full-block acceptance increases dramatically, indicating that the proposed prefix-conditioned tree construction is more effective at identifying globally consistent candidate paths rather than locally confident but incompatible drafts. This distributional shift directly explains the throughput improvements observed in the main results.

\section{Related Work}
\label{app:related_work}
\subsection{Speculative Decoding}

Speculative decoding~\citep{leviathan2023fastinferencetransformersspeculative, chen2023acceleratinglargelanguagemodel} accelerates AR models by drafting tokens with a fast proposer and verifying them in parallel with the target model, provably preserving the target distribution. This guarantee relies on the target model possessing a well-trained verify distribution. A line of follow-up work eliminates the external drafter or improves draft quality: Medusa~\citep{cai2024medusasimplellminference} augments the base LLM with multiple prediction heads and uses tree attention for parallel verification; the EAGLE family~\citep{li2025eaglespeculativesamplingrequires, li2024eagle2fasterinferencelanguage, li2025eagle3scalinginferenceacceleration} exploits feature-level context from the frozen target model, with EAGLE-1 predicting future hidden states to boost acceptance, EAGLE-2 introducing adaptive drafting trees, and EAGLE-3 refining training objectives to scale speedups; and SpecInfer~\citep{Miao_2024} proposes tree-structured verification. Multi-token prediction (MTP)~\citep{samragh2025llmknowsfutureuncovering} trains models to predict multiple future tokens simultaneously. Despite these advances, most existing methods rely on autoregressive drafting, which remains inherently sequential and limits attainable speedups.

\subsection{Tree Construction in Speculative Decoding}

A complementary line of work focuses on how draft tokens are organized for parallel verification. Medusa~\citep{cai2024medusasimplellminference} popularized tree-structured drafting by packing multiple candidate continuations into a single token tree and verifying them in one forward pass via a topology-aware attention mask. However, this tree mask is hand-crafted and static. SpecInfer~\citep{Miao_2024} introduced token-tree verification with provable distribution preservation. Subsequent work moves beyond fixed shapes: Sequoia~\citep{chen2025sequoiascalablerobusthardwareaware} formulates tree construction as a dynamic-programming problem, jointly optimizing the tree topology and a hardware-aware tree size for a given accelerator. SpecExec~\citep{svirschevski2024specexecmassivelyparallelspeculative} pushes draft trees to massively parallel sizes (thousands of nodes) to amortize the cost of offloaded weights on consumer hardware based on Best First Search like tree construction algorihtm. EAGLE-2~\citep{li2024eagle2fasterinferencelanguage} observes that draft acceptance is context-dependent and uses the calibrated confidence scores of the EAGLE drafter to construct a dynamic draft tree per step via beam search like tree expension. OPT-Tree~\citep{wang2025opttreespeculativedecodingadaptive} formalizes the objective as maximizing the expected acceptance length and searches for the adaptive tree structure that attains it under a node budget. These methods consistently show that careful tree drafting contributes a substantial fraction of end-to-end speedup. However, they all assume an autoregressive drafter that produces tokens sequentially with conditional probabilities along each branch, and extending tree construction to diffusion-based drafters, whose draft tokens are produced in parallel remains an open problem that our work directly addresses.

\subsection{Diffusion-based Speculative Decoding}

Recent work explores using diffusion models as drafters within speculative decoding, combining the parallelism of diffusion drafting with the quality guarantee of AR verification. TiDAR~\citep{liu2025tidarthinkdiffusiontalk} jointly trains diffusion and autoregressive objectives in a sequence-level hybrid, enabling parallel ``thinking'' via diffusion and sequential ``talking'' via autoregressive decoding, though final generation quality is not yet lossless. DiffuSpec~\citep{li2025diffuspecunlockingdiffusionlanguage} and SpecDiff-2~\citep{sandler2025specdiff2scalingdiffusiondrafter} employ large pretrained dLLMs as speculative drafters, with inference-time search or train--test alignment to improve acceptance. However, these approaches primarily focus on improving the quality or alignment of the diffusion drafter itself, often relying on massive drafters (\textit{e.g.}, 7B parameters), which incur substantial memory and latency overhead. While they achieve long acceptance lengths, the high drafting cost can offset the practical speedups in real-world serving scenarios. In contrast, PRESTO studies how to more effectively exploit the rich multi-position candidate space produced by diffusion drafting through prefix-aware tree construction and priority-based expansion, making it complementary to stronger diffusion drafters and existing diffusion-based speculative decoding systems. DFlash~\citep{chen2026dflashblockdiffusionflash} proposes to employ a lightweight block-diffusion drafter conditioned on context features extracted from the target model, generating an entire block of draft tokens in a single forward pass and reporting over $6\times$ lossless acceleration. Similarly, DART~\citep{liu2026dartdiffusioninspiredspeculativedecoding} performs parallel logit prediction over multiple masked positions and assembles drafts via N-gram-guided tree pruning. However, its tree construction relies on a fixed, hand-tuned scoring rule with hard-coded hyperparameters, which leaves no principled way to adapt the tree shape to different budgets, drafters, or target models without re-tuning. In contrast, PRESTO formulates tree construction as priority-based expansion under a unified path score, yielding a single algorithmic framework that naturally generalizes across different tree budgets and diffusion drafter designs.

\section{Conclusion}
In this work, we propose PRESTO, a principled framework for tree-based speculative decoding with diffusion language models. We identify a fundamental mismatch between diffusion draft scoring and prefix-based autoregressive verification: while diffusion probabilities provide strong marginal plausibility signals, they are inherently prefix-blind and therefore insufficient for reliable path ranking in tree-based drafting. To address this issue, we introduce a prefix-aligned scoring mechanism together with a priority-based tree search strategy, enabling effective exploration of high-quality candidate paths during diffusion drafting. 

\section{Limitations}

PRESTO requires a tractable prefix-aligned signal $\rho_d$ to complement the diffusion drafter's prefix-blind marginals. In this work, we employ a lightweight n-gram model, which is empirically effective, but only captures short-range lexical compatibility. Exploring richer prefix-conditioned signals while preserving low overhead remains an important future direction. We hope this motivates further efforts to inject prefix-aligned signals into diffusion marginals. 

\noindent We additionally explored an entropy-adaptive variant of $\lambda_d$, but observed only marginal improvements over a fixed coefficient, suggesting that more sophisticated adaptation strategies may require learned or target-aware scheduling mechanisms.

\noindent Finally, we primarily evaluate PRESTO under single-request decoding settings using the native PyTorch implementation except NLD at linear self speculation decoding mode. Extending PRESTO to production-scale serving frameworks such as vLLM and SGLang, together with characterizing its behavior under larger batch sizes, longer contexts, and highly optimized serving systems, remains meaningful future work.

\newpage
{
  \small
  \bibliographystyle{unsrt}
  \bibliography{main_tech_report}
}

\newpage
\appendix
\section{Prefix-Mass Decomposition of the Tree Objective}
\label{appendix:decomposition}

Let $\mathcal{V}$ denote the token vocabulary and let 
$\mathcal{T}$ be a prefix-closed tree whose nodes are finite token 
sequences. Define
\[
\mathcal{T}^{+} := \mathcal{T}\setminus\{\emptyset\}
\]
as the set of non-root nodes. For each node 
$u=(u_1,\ldots,u_{|u|})\in\mathcal{T}^{+}$, let $|u|$ denote its depth.

For a sampled path 
\[
P=(x_1,\ldots,x_k)\sim\tilde p,
\]
define the length-$d$ prefix
\[
P_{\le d}:=(x_1,\ldots,x_d).
\]
The accepted-prefix length under tree $\mathcal{T}$ is
\[
\alpha_{\mathcal T}(P)
:=
\max\{d : P_{\le d}\in\mathcal T\}.
\]

Since $\mathcal T$ is prefix-closed, the event 
$P_{\le d}\in\mathcal T$ implies 
$P_{\le j}\in\mathcal T$ for all $j\le d$. Hence,
\[
\alpha_{\mathcal T}(P)
=
\sum_{d=1}^{k}
\mathbf 1\{P_{\le d}\in\mathcal T\}.
\]
Equivalently,
\[
\alpha_{\mathcal T}(P)
=
\sum_{u\in\mathcal T^{+}}
\mathbf 1\{P_{\le |u|}=u\}.
\]

Taking expectation with respect to 
$P\sim\tilde p$ and applying linearity of expectation,
\[
\mathbb E_{P\sim\tilde p}[\alpha_{\mathcal T}(P)]
=
\sum_{u\in\mathcal T^{+}}
\Pr_{P\sim\tilde p}(P_{\le |u|}=u).
\]

Define the surrogate prefix mass
\[
\tilde p(u)
:=
\Pr_{P\sim\tilde p}(P_{\le |u|}=u).
\]
Then
\[
\mathbb E_{P\sim\tilde p}[\alpha_{\mathcal T}(P)]
=
\sum_{u\in\mathcal T^{+}}
\tilde p(u).
\]

If the root node is excluded by convention, then 
$\mathcal T^{+}=\mathcal T$, yielding Eq.~\eqref{eq:surrogate-objective}.

\newpage
\section{KL-Regularized Derivation of the Prefix-Aligned Surrogate}
\label{app:kl-derivation}

We consider the variational objective
\begin{equation}
\min_{p \in \Delta(\mathcal V)}
\;
\mathrm{KL}\!\left(
p \,\|\, q_d(\cdot)
\right)
-
\lambda_d
\mathbb E_{t\sim p}
\bigl[
\log \rho_d(t\mid c_d)
\bigr],
\label{eq:appendix-kl}
\end{equation}
where $q_d$ denotes the diffusion drafter marginal, 
$\rho_d(\cdot\mid c_d)$ is a prefix-conditioned compatibility signal, 
and $\Delta(\mathcal V)$ denotes the probability simplex over the vocabulary. We assume $q_d(t)>0$ and $\rho_d(t\mid c_d)>0$ on the candidate support.

\noindent Expanding the KL divergence,
\[
\mathrm{KL}(p\|q_d)
=
\sum_t p(t)\log\frac{p(t)}{q_d(t)},
\]
the objective becomes

\[
\mathcal L(p)
=
\sum_t
p(t)\log\frac{p(t)}{q_d(t)}
-
\lambda_d
\sum_t
p(t)\log\rho_d(t\mid c_d).
\]

\noindent Including the normalization constraint 
$\sum_t p(t)=1$ with Lagrange multiplier $\mu$, we obtain
\[
\mathcal J(p)
=
\sum_t
p(t)\log\frac{p(t)}{q_d(t)}
-
\lambda_d
\sum_t
p(t)\log\rho_d(t\mid c_d)
+
\mu\left(
\sum_t p(t)-1
\right).
\]

\noindent Taking derivatives with respect to $p(t)$,
\[
\frac{\partial\mathcal J}{\partial p(t)}
=
\log p(t)
-
\log q_d(t)
+
1
-
\lambda_d\log\rho_d(t\mid c_d)
+
\mu.
\]

\noindent Setting the derivative to zero yields
\[
\log p^\star_d(t)
=
\log q_d(t)
+
\lambda_d\log\rho_d(t\mid c_d)
+
C,
\]
where $C$ absorbs constants independent of $t$. Exponentiating both sides,
\[
p^\star_d(t)
\propto
q_d(t)\,
\rho_d(t\mid c_d)^{\lambda_d}.
\]
Equivalently,
\[
p^\star_d(t\mid c_d)
=
\frac{
q_d(t)\rho_d(t\mid c_d)^{\lambda_d}
}{
Z_d(c_d)
},
\qquad
Z_d(c_d)
=
\sum_{t'\in\mathcal V}
q_d(t')\rho_d(t'\mid c_d)^{\lambda_d}.
\]

\noindent Thus, the optimal surrogate takes a product-of-experts form, combining the calibrated marginal signal $q_d$ with a prefix-conditioned correction $\rho_d$.

\newpage
\section{PRESTO on Hybrid dLLMs at Quadratic Self Speculation Decoding}
\label{app:self_speculative}

We provide a detailed walk-through of how PRESTO is instantiated on Hybrid dLLMs at quadratic self speculation decoding mode, corresponding to
Figure~\ref{fig:pipeline_speculative_dllm}. Unlike the standard speculative decoding
setup where a small drafter proposes tokens for a larger verifier,
self-speculative dLLMs use \emph{the same dLLM} as both drafter and
verifier: a single forward pass simultaneously verifies previously
drafted tokens and speculates new ones at trailing mask positions.
PRESTO turns this single-model pipeline into a tree-based drafter
without any additional model.

At the start of round $r$, the input sequence consists of three
contiguous regions:
\begin{itemize}
    \item a \textbf{committed prefix} of accepted tokens from previous
    rounds (green tokens with \checkmark{} in Figure~\ref{fig:pipeline_speculative_dllm},
    e.g.\ \texttt{C, D});
    \item a \textbf{verification region} containing the draft tokens
    proposed in round $r-1$ (e.g.\ \texttt{E, F, G}), some of which will
    be accepted and some rejected;
    \item a \textbf{speculation region} of mask tokens (\texttt{M})
    appended to the right, at which the dLLM will produce new draft
    candidates.
\end{itemize}
A single forward pass over this composite input yields outputs that
serve both roles below.

\paragraph{(I) AR verification of the previous draft tree.}
On the verification region, PRESTO uses the exact KV cache of the
committed prefix together with an \emph{AR tree attention mask} that
restricts each draft token to attend only to its ancestors in the
previous round's draft tree. This lets a single forward verify all
paths of the tree in parallel: each path is checked left-to-right
against the model's predictions, and the longest accepted prefix
across paths is committed (\texttt{D}$\to$\texttt{F} in
Figure~\ref{fig:pipeline_speculative_dllm}, where \texttt{E} on the sibling path and
\texttt{G} on the same path are rejected). The position immediately
after the last accepted token is filled with the model's own
prediction (\texttt{F'}, \texttt{G'}, \texttt{H'} in
Figure~\ref{fig:pipeline_speculative_dllm}), and the accepted tokens become part of
the committed prefix for round $r+1$.

\paragraph{(II) Diffusion-based speculation along the accepted path.}The speculation region is structured as a set of mask slots attached
to each draft token from the previous round, mirroring the previous
round's draft tree (Figure~\ref{fig:pipeline_speculative_dllm} shows two candidate
paths under verification, \texttt{D}$\to$\texttt{E} and
\texttt{D}$\to$\texttt{F}$\to$\texttt{G}). The dLLM denoises all of
these mask positions in parallel within the same forward, but only
the mask slots hanging off the \emph{accepted} path are used for the
next round's drafting. In Figure~\ref{fig:pipeline_speculative_dllm}, \texttt{F} is
accepted and \texttt{G} is rejected, so PRESTO reads top-$b$
candidates from the mask slots following \texttt{F} ($b = 2$ in the
figure, e.g.\ $\{H_1, H_2\}$ at the first such position,
$\{I_1, I_2\}$ at the next, and so on). The mask slots attached to
rejected branches are produced by the same forward but discarded
(marked ``Unused'' in Figure~\ref{fig:pipeline_speculative_dllm}), since their
predictions were conditioned on a context that is no longer part of
the committed prefix.

\paragraph{(III) Draft tree construction.}
The per-position top-$b$ candidates are assembled into a draft tree
following the priority-based expansion of
Section~\ref{sec:method_search}. The root corresponds to the
first speculation position, and at depth $d$ each node has up to $b$
children drawn from the top-$b$ candidates at depth $d+1$. Each child
inherits its parent's cumulative path score $S$ and adds the local
token score $s_{d+1, t}(c)$, so scores are computed incrementally with
no redundant work. The frontier is expanded under beam search with
global retention until the tree contains $B$ nodes, after which the
top-$B$ paths under $S$ are kept (Algorithm~\ref{alg:beam}). Crucially,
the candidate logits at every depth come from the \emph{same} forward
pass that performed verification, so tree construction itself adds no
additional dLLM forwards.

\paragraph{(IV) Flattening for the next forward pass.}
The draft tree is then linearised into the input sequence for round
$r+1$ (bottom of Figure~\ref{fig:pipeline_speculative_dllm}). Each draft token is
followed by a block of mask tokens that will host the next round's
speculation, while the tree's parent--child structure is encoded in
the AR tree attention mask used during verification. The result is a
single input on which the next forward pass again performs
verification and speculation jointly, completing the cycle.

\newpage
\section{End-to-End Throughput Results for dFlash}
\label{app:throughput_dflash}
\begin{table}[h]
\centering
\small
\caption{Average absolute decoding throughput (tokens/s) on Qwen3 models with thinking mode disabled. Speedup is measured against autoregressive decoding on torch implementation with single batch size.}
\label{tab:absolute-throughput-dflash}
\begin{tabular}{ccccc}
\toprule
Model & Temp. & dFlash & \name & Gain \\
\midrule
\multirow{2}{*}{Qwen3-4B}
& $T=0$ & 211 (4.9$\times$) & \textbf{314} (7.3$\times$) & 1.49$\times$ \\
& $T=1$ & 168 (3.9$\times$) & \textbf{258} (6.0$\times$) & 1.54$\times$ \\
\midrule
\multirow{2}{*}{Qwen3-8B}
& $T=0$ & 173 (4.8$\times$) & \textbf{263} (7.3$\times$) & 1.52$\times$ \\
& $T=1$ & 137 (3.8$\times$) & \textbf{216} (6.0$\times$) & 1.58$\times$ \\
\bottomrule
\end{tabular}
\end{table}

% \newpage
\section{End-to-End Throughput Results for Nemotron-Labs-Diffusion}
\label{app:throughput_nld}

\begin{table*}[hth]
\centering
\small
\setlength{\tabcolsep}{2pt}
\caption{Absolute decoding throughput (tokens/s), average acceptance length ($\tau$), and average throughput improvement on Nemotron-Labs-Diffusion-8B under linear self-speculative decoding with a maximum generation length of 2048 tokens based on SGLang implementation.}
\label{tab:throughput_nld}
\resizebox{\linewidth}{!}{%
\begin{tabular}{
c l
@{\hspace{0.3em}} cc cc cc cc
@{\hspace{0.3em}} cc cc cc cc
@{\hspace{0.3em}} cc cc
@{\hspace{0.3em}} cc
}
\toprule
\multirow{2}{*}{Model} & \multirow{2}{*}{Method}
& \multicolumn{8}{c@{\hspace{1.2em}}}{\sc{Math}}
& \multicolumn{8}{c@{\hspace{1.2em}}}{\sc{Code}}
& \multicolumn{4}{c@{\hspace{1.2em}}}{\sc{Chat}}
& \multicolumn{2}{c}{\textit{Avg.}} \\
\cmidrule(lr){3-10} \cmidrule(lr){11-18} \cmidrule(lr){19-22} \cmidrule(lr){23-24}

& & \multicolumn{2}{c}{GSM8K}
& \multicolumn{2}{c}{MATH-500}
& \multicolumn{2}{c}{AIME24}
& \multicolumn{2}{c@{\hspace{1.2em}}}{AIME25}
& \multicolumn{2}{c}{HumanEval}
& \multicolumn{2}{c}{MBPP}
& \multicolumn{2}{c}{LCB}
& \multicolumn{2}{c@{\hspace{1.2em}}}{SWE-Bench}
& \multicolumn{2}{c}{MT-Bench}
& \multicolumn{2}{c}{Alpaca}
& \multicolumn{2}{c}{\textit{Avg.}} \\
\midrule

\multicolumn{2}{c}{Temperature = 0}
& \tabsmall Thr. & \tabsmall $\tau$
& \tabsmall Thr. & \tabsmall $\tau$
& \tabsmall Thr. & \tabsmall $\tau$
& \tabsmall Thr. & \tabsmall $\tau$
& \tabsmall Thr. & \tabsmall $\tau$
& \tabsmall Thr. & \tabsmall $\tau$
& \tabsmall Thr. & \tabsmall $\tau$
& \tabsmall Thr. & \tabsmall $\tau$
& \tabsmall Thr. & \tabsmall $\tau$
& \tabsmall Thr. & \tabsmall $\tau$
& \tabsmall Thr. & \tabsmall $\tau$ \\
\midrule

\multirow{4}{*}{8B}
& AR
& 151 & 1.0
& 139 & 1.0
& 130 & 1.0
& 127 & 1.0
& 142 & 1.0
& 148 & 1.0
& 126 & 1.0
& 125 & 1.0
& 142 & 1.0
& 145 & 1.0
& 136 & 1.0 \\

& NLD
& 868 & 10.8
& 901 & 12.2
& 746 & 11.0
& 700 & 10.5
& 710 & 9.3
& 659 & 8.2
& 593 & 9.0
& 471 & 6.8
& 402 & 5.2
& 363 & 4.6
& 641 & 8.7 \\

& NLD w/ \name
& 908 & 11.9
& 933 & 13.2
& 792 & 12.2
& 731 & 11.5
& 738 & 10.3
& 681 & 9.1
& 623 & 10.0
& 506 & 7.9
& 460 & 6.5
& 420 & 5.8
& 679 & 9.9 \\

\cmidrule(lr){2-24}
\rowcolor[rgb]{0.88,0.96,0.84}
& \textbf{$\times$ Gain}
& \textbf{1.05$\times$} &
& \textbf{1.03$\times$} &
& \textbf{1.06$\times$} &
& \textbf{1.04$\times$} &
& \textbf{1.04$\times$} &
& \textbf{1.03$\times$} &
& \textbf{1.05$\times$} &
& \textbf{1.07$\times$} &
& \textbf{1.14$\times$} &
& \textbf{1.16$\times$} &
& \textbf{1.06$\times$} & \\

\midrule

\multicolumn{2}{c}{Temperature = 1}
& \tabsmall Thr. & \tabsmall $\tau$
& \tabsmall Thr. & \tabsmall $\tau$
& \tabsmall Thr. & \tabsmall $\tau$
& \tabsmall Thr. & \tabsmall $\tau$
& \tabsmall Thr. & \tabsmall $\tau$
& \tabsmall Thr. & \tabsmall $\tau$
& \tabsmall Thr. & \tabsmall $\tau$
& \tabsmall Thr. & \tabsmall $\tau$
& \tabsmall Thr. & \tabsmall $\tau$
& \tabsmall Thr. & \tabsmall $\tau$
& \tabsmall Thr. & \tabsmall $\tau$ \\
\midrule

\multirow{4}{*}{8B}
& AR
& 150.8 & 1.0
& 138.4 & 1.0
& 128.8 & 1.0
& 125.4 & 1.0
& 141.5 & 1.0
& 147.1 & 1.0
& 125.4 & 1.0
& 124.2 & 1.0
& 141.5 & 1.0
& 144.5 & 1.0
& 136.7 & 1.0 \\

& NLD
& 431.6 & 5.6
& 401.1 & 5.7
& 315.9 & 5.0
& 330.0 & 4.9
& 322.0 & 4.6
& 310.4 & 4.2
& 250.6 & 4.0
& 175.8 & 2.9
& 194.2 & 2.7
& 176.0 & 2.3
& 290.7 & 4.2 \\

& NLD w/ \name
& 470.6 & 6.1
& 452.6 & 7.1
& 378.2 & 5.7
& 357.7 & 6.4
& 376.1 & 5.5
& 376.0 & 5.0
& 284.3 & 4.8
& 221.6 & 3.6
& 252.0 & 3.5
& 213.0 & 3.2
& 338.2 & 5.1 \\

\cmidrule(lr){2-24}
\rowcolor[rgb]{0.88,0.96,0.84}
& \textbf{$\times$ Gain}
& \textbf{1.09$\times$} &
& \textbf{1.13$\times$} &
& \textbf{1.20$\times$} &
& \textbf{1.08$\times$} &
& \textbf{1.17$\times$} &
& \textbf{1.21$\times$} &
& \textbf{1.13$\times$} &
& \textbf{1.26$\times$} &
& \textbf{1.30$\times$} &
& \textbf{1.21$\times$} &
& \textbf{1.16$\times$} &\\

\bottomrule
\end{tabular}%
}
\end{table*}

\newpage
\section{Entropy-adaptive $\lambda_{d}$}
\label{app:entropy-adaptive}
In addition to a fixed interpolation coefficient $\lambda_d$, we
explored an entropy-adaptive variant that scales the prefix
correction by the uncertainty of the diffusion drafter. Let
\[
H_d = -\sum_{w \in \mathcal{V}} q_d(w) \log q_d(w)
\]
denote the entropy of the diffusion marginal at depth $d$. We define
\[
\lambda_d(H_d) = \beta \cdot \frac{H_d}{H_d + C},
\]
where $\beta > 0$ controls the maximum correction strength and $C>0$
is a stabilising constant, and substitute this into the token-level
score from Section 4:
\[
s_{d,t}(c_d) = \log q_d(t) + \lambda_d(H_d)\, \log \rho_d(t \mid c_d).
\]
Under this schedule, low-entropy (high-confidence) positions assign
small weight to the prefix correction and rely primarily on the
calibrated marginal $q_d$, while high-entropy positions up-weight the
prefix-conditioned signal $\rho_d$. The fixed-$\lambda_d$ variant
used in our main experiments is recovered by replacing
$\lambda_d(H_d)$ with a constant.

\end{document}